\documentclass[manuscript, screen]{acmart}

\AtBeginDocument{%
  }

\setcopyright{acmlicensed}
\copyrightyear{2018}
\acmYear{2018}
\acmDOI{XXXXXXX.XXXXXXX}

\acmConference[Conference acronym 'XX]{Make sure to enter the correct
  conference title from your rights confirmation emai}{June 03--05,
  2018}{Woodstock, NY}
\acmISBN{978-1-4503-XXXX-X/18/06}

\usepackage{multirow}
\usepackage{longtable}
\usepackage{makecell}
\usepackage{float}
\usepackage{subcaption}
\usepackage{wrapfig}
\usepackage{dsfont}
\usepackage{amsmath}
\usepackage{threeparttable}

\newcommand{\mC}{\mathcal{C}}
\newcommand{\mX}{\mathcal{X}}
\newcommand{\mZ}{\mathcal{Z}}
\newcommand{\mF}{\mathcal{F}}
\newcommand{\mL}{\mathcal{L}}
\newcommand{\mG}{\mathcal{G}}
\newcommand{\mD}{\mathcal{D}}
\newcommand{\mE}{\mathcal{E}}
\newcommand{\mN}{\mathcal{N}}

\begin{document}

\title{Bridging the Gap: A Decade Review of Time-Series Clustering Methods}



\author{John Paparrizos}
\affiliation{%
  \institution{The Ohio State University}
  \city{Columbus}
  \country{USA}}
\email{paparrizos.1@osu.edu}

\author{Fan Yang}
\affiliation{%
  \institution{The Ohio State University}
  \city{Columbus}
  \country{USA}}
\email{yang.7007@osu.edu}

\author{Haojun Li}
\affiliation{%
  \institution{The Ohio State University}
  \city{Columbus}
  \country{USA}}
\email{li.14118@.osu.edu}

\renewcommand{\shortauthors}{Paparrizos et al.}

\begin{abstract}
  Time series, as one of the most fundamental representations of sequential data, has been extensively studied across diverse disciplines, including computer science, biology, geology, astronomy, and environmental sciences. The advent of advanced sensing, storage, and networking technologies has resulted in high-dimensional time-series data, however, posing significant challenges for analyzing latent structures over extended temporal scales. Time-series clustering, an established unsupervised learning strategy that groups similar time series together, helps unveil hidden patterns in these complex datasets. In this survey, we trace the evolution of time-series clustering methods from classical approaches to recent advances in neural networks. While previous surveys have focused on specific methodological categories, we bridge the gap between traditional clustering methods and emerging deep learning-based algorithms, presenting a comprehensive, unified taxonomy for this research area. This survey highlights key developments and provides insights to guide future research in time-series clustering.
  
\end{abstract}





\received{20 February 2007}
\received[revised]{12 March 2009}
\received[accepted]{5 June 2009}

\maketitle

\section{Introduction}
Time series, an ordered sequence of real-valued data, has been widely acknowledged as one of the most basic data formats. With the development of technologies in sensing, storage, networking, and data mining, massive raw data could be obtained, stored, and processed on the fly \cite{paparrizos2021vergedb, paparrizos2018fast}. Due to the advantage of chronological representation, we could see the application of time series in almost every scientific field or industry \cite{paparrizos2016detecting,paparrizos2016screening,goel2016social,mckeown2016predicting,jeung2010effective, liu2023amir,krishnan2019artificial,YANG2024acloserlook,li2023towards-noise}, including but not limited to: Environment \cite{iglesias2013analysis,kovsmelj1990cross,steinbach2003discovery}, Biology \cite{subhani2010multiple,wismuller2002cluster,zupko2023modeling}, Finance \cite{fu2001pattern,aghabozorgi2014stock,stetco2013fuzzy,guam2007cluster}, Psychology \cite{kurbalija2012time}, Artificial Intelligence \cite{tran2002fuzzy}. 
However, in the information era, increasing data sizes that contain thousands or even millions of dimensions have become increasingly common. 
This has introduced a new layer of complexity, requiring efficient adaptive solutions \cite{dziedzic2019band,jiang2020pids,jiang2021good,liu2021decomposed,liu2024adaedge} and presenting challenges in analyzing the underlying relationships between time series in various tasks, including indexing \cite{shieh2009isax, paparrizos2023querying, paparrizos2020debunking, paparrizos2022fast,paparrizos2023accelerating,d2024beyond}, anomaly detection \cite{sylligardos2023choose, boniol2021sand, boniol2021sandinaction, boniol2023new, boniol2022theseus, paparrizos2022tsb,paparrizos2022volume,liu2024elephant,boniol2024adecimo,boniol2024interactive,liu2024time}, clustering \cite{paparrizos2015k, paparrizos2017fast, paparrizos2023odyssey, bariya2021k}, classification \cite{shaw1992using, paparrizos2019grail}, and forecasting \cite{pavlidis2006financial}.

\begin{figure}[!t]
	\centering
	\includegraphics[width=\textwidth]{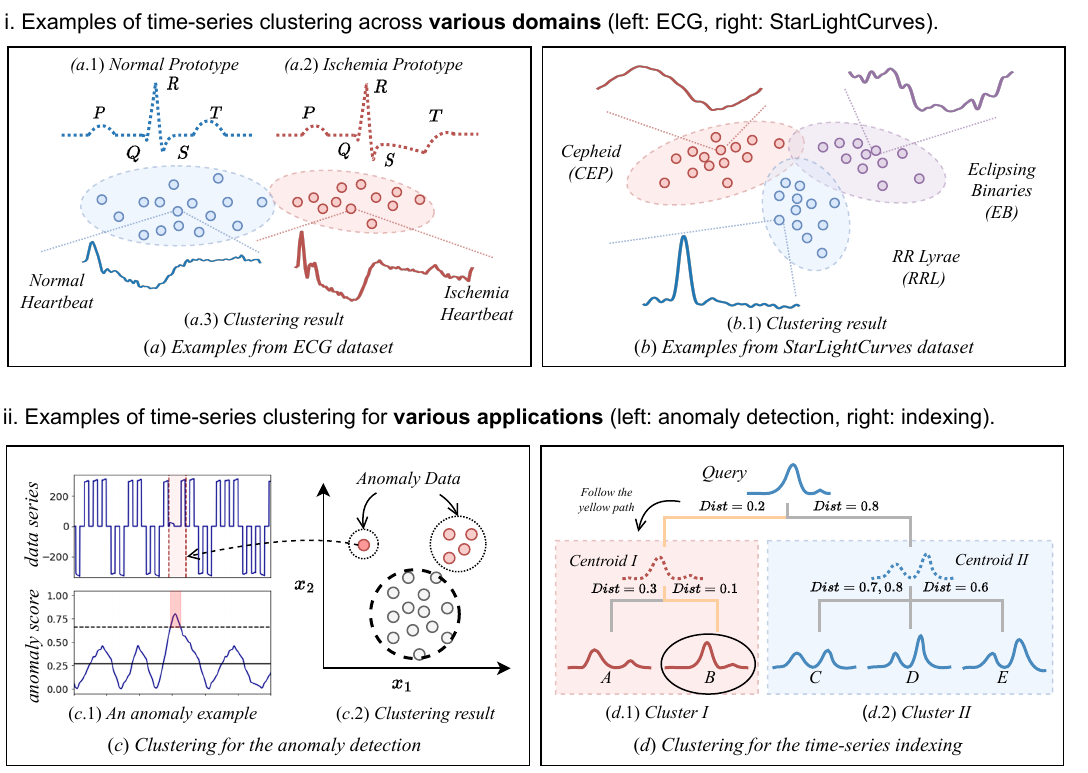}
	\caption{Examples of time-series clustering across different domains and applications.}
	\label{pic:intro}
\end{figure}

Clustering has been one of the earliest concepts developed in the field of unsupervised machine learning, which is shown to be one of the most efficient tools to help unveil the latent structure from the raw data. The main goal of clustering is to find a partition of various given objects, in which the similarity within each group is maximized, and minimized between groups. In other words, a group of ``similar" data samples is viewed as a cluster. Consequently, the characteristics within each cluster of objects consist of the \textit{pattern} in the dataset, i.e., common features shared within the cluster. For example, in computer vision tasks, the pattern can be a certain type of edge or color for similar objects \cite{lowe2004distinctive}, while in the time-series domain, the position of rapid growth or decline across time steps may have practical meaning. On the one hand, data samples sharing the same pattern would have a small distance and thus be partitioned into the same cluster. On the other hand, a good clustering strategy would in turn facilitate the search for these representative features. Figure \ref{pic:intro} depicts the examples of time-series clustering across different domains and applications.  

As one of the most well-known clustering methods, the k-Means algorithm \cite{macqueen1967some} provides an expectation–maximization (EM) based strategy to search the medoids and clustering assignments based on Euclidean Distance for each iteration. Its good performance has enabled the application in different fields such as electrical engineering, computer science, biology, and finance.
However, traditional methods such as k-Means have suffered from significant performance degradation in the time-series domain. 
Traditional measurements in Euclidean space have proven inefficient for addressing the variety of distortions in time-series data, including shifting, scaling, and occlusion \cite{paparrizos2015k},
which are common in real-world scenarios. To solve this problem, many methods have been proposed, which can offer invariances to the inherent distortions and robustness to noise or outliers. For example, Dynamic Time Warping (DTW) \cite{chu2002iterative} proposes an elastic measure to deal with the many-to-many alignment issue and finds the optimal warping path that minimizes the total distance between 2 sequences. In order to further reduce the computation cost, k-Shape \cite{paparrizos2015k} introduces (i) shape-based distance (SBD) with time complexity of $\mathcal{O}(n\log(n))$ and (ii) a novel centroid computation derived from an optimization problem, which achieves a significant improvement in clustering tasks \cite{paparrizos2015k,paparrizos2023odyssey}. In recent years, numerous clustering methods have emerged from the deep learning era, capturing significant attention and interest. Various unsupervised learning strategies, like contrastive learning, enable neural networks to generate representative features in a reduced dimension space, which significantly alleviates the pressure of the downstream tasks such as dissimilarity measure and centroid computation. Leveraging parallel computing strategies and advanced GPU resources, a clustering model can be trained and deployed in significantly less time.

Past reviews \cite{aghabozorgi2015time,liao2005clustering} have explored time-series clustering algorithms proposed in decades and provided insights from various perspectives of views, i.e., data representation, dissimilarity measure, clustering methods, and evaluation metrics. However, many survey papers, as mentioned above, either only discuss the conventional time-series clustering before the deep learning era \cite{aghabozorgi2015time,liao2005clustering} or mainly focus on end-to-end deep representation learning \cite{lafabregue2021end}. \cite{alqahtani2021deep} reviews the conventional time-series clustering works and prior deep clustering methods. However, it mainly focuses on the case study in the context of biological time-series clustering without a comprehensive study on both sides as mentioned. To the best of our knowledge, this work is the first attempt to build a bridge between the conventional time-series clustering methods and the deep learning-based models, providing a novel and comprehensive taxonomy for various time-series clustering in each category. We anticipate that this work will provide valuable insights for next-generation clustering algorithm designs.

\section{Time-series Clustering Overview}
In this section, we first introduce the definition of time-series data, and the difference between univariate and multivariate time series. Then we present the problem formulation of time-series clustering and the general pipeline of the clustering process, which motivates the newly proposed taxonomy in the following section.

\subsection{On the Definition of Time-series Data}
\label{sec: 2.1}


Time-series data can be categorized based on their major characteristics. From the perspective of dimensionality, they can be classified into three main types: univariate, multivariate, and tensor fields. Additionally, depending on the sampling strategies employed during data acquisition, time-series data can be either regular or irregular. In the following sections, we will explore each of these categories in detail. The formal definition of time-series data is provided below.

\begin{definition} [Time series]
A time series $x_i$ is defined as a sequence of observations with $T>1$ time steps $x_i = \{x_{i1}, x_{i2}, \dots, x_{iT}\}$, where $x_{it} \in \mathbb{R}^{d}$. Based on the dimension $d$, each observation $x_{it}$ at time $t$ can be a real-valued number ($d=1$) or a vector ($d>1$).
\end{definition}

\begin{definition} [Subsequence]

Given a time series $x_i = \{x_{i1}, x_{i2}, \dots, x_{iT}\}$, where $x_{it} \in \mathbb{R}^{d}$, a subsequence $C_i$ is defined as a sequence of consecutive time steps obtained from $x_i$: $C_i = \{x_{im}, x_{i(m+1)}, \cdots, x_{i(m+L-1)}\}$, where $L$ is the length of the subsequence, $1\leq m\leq T+1-L$.
\end{definition}

\begin{definition} [Sliding Windows]
\label{def: sliding windows}
Given a Time series $x_i = \{x_{i1}, x_{i2}, \dots, x_{iT}\}$, where $x_{it} \in \mathbb{R}^{d}$, sliding windows are defined as a set of subsequences extracted by sliding a ``window" of length $L$ across the current time series $x_i$, with a stride size $s$ ($1 \leq L \leq T - 1$ and $1 \leq s \leq T-L$). The size of the sliding windows matrix is $(\lfloor \frac{T-L}{s} \rfloor+1, L)$.
\end{definition}

\subsubsection{Univariate versus Multivariate}

\begin{figure}[!t]
\centering
\includegraphics[width=\linewidth]{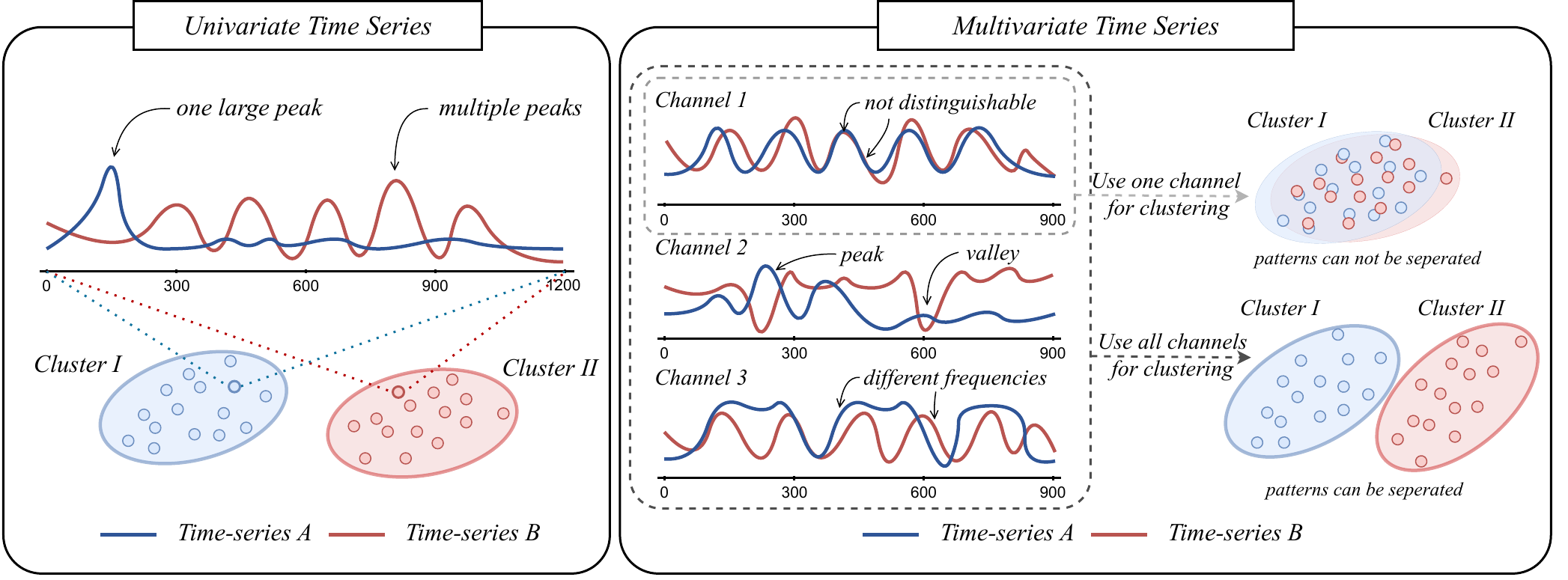}
\caption{Examples of univariate time series (left) and multivariate time series (right). Compared with univariate time-series clustering, multivariate time-series clustering needs to consider time steps from all channels.}
\label{pic:TSClustering_uni_multi}
\end{figure}

From the discussion above, we could clearly categorize the univariate and multivariate time series based on the dimension of each observation across time steps. Univariate time series (UTS) represents an ordered sequence of real-valued data in one dimension ($d=1$), e.g., ECG200 in UCR dataset \cite{UCRArchive} consists of 200 ECG recordings of a single patient, each indicating the changes of electric activity during one heartbeat. Compared with univariate time series, the multivariate time series (MTS) contains observations of more than one dimension ($d>1$). For example, PenDigits in UEA dataset \cite{bagnall2018uea} records the movement of both $x$ and $y$ coordinates when writers draw digits between 0 and 9. Compared with UTS, Methods designed for MTS need to address the dependencies across multiple channels, which adds complexity and poses additional challenges to the clustering task in MTS scenarios.

Figure \ref{pic:TSClustering_uni_multi} presents examples of clustering for both univariate and multivariate time series, highlighting the importance of considering time steps from all channels in the development of algorithms for multivariate time-series clustering. The left side of this figure illustrates that we can directly categorize two univariate time series into two distinct clusters by observing the difference in the number of peaks between them. On top of that, the right side of the figure shows that clustering two multivariate time series by using the rule that only considers a single channel, a univariate time series, is inadequate. Rather, a MTS clustering algorithm should incorporate the impact of all channels to ensure accuracy.



\subsection{On the Definition of Time-series Clustering}
\label{sec: 2.2}

Clustering, as one of the earliest concepts in the machine learning field, has been widely applied in the time-series domain. The overall goal of clustering is to find a solution to group different data samples in a way that, the distances, or the dissimilarity measurements, within each group are minimized, and the distances between each group are maximized. This clustering procedure not only finds a special partition way for a whole dataset, but also provides valuable insights for understanding the latent structure of the data and strongly facilitates downstream tasks such as time-series classification, segmentation, anomaly detection, etc. The detailed definition is provided below.

\begin{definition}
    \cite{aghabozorgi2015time} Given a dataset of $N$ time-series data $\mathcal{X}=\{x_1, x_2, \cdots, x_N \}$, where $x_i\in \mathbb{R}^{d\times T}$, the process of time-series clustering is to partition $\mathcal{X}$ in to $K$ clusters $\mathcal{C}=\{c_1, c_2, \cdots ,c_K\}$, $K<N$. In general, homogenous time-series data that share similar characteristics are grouped together based on a pre-defined dissimilarity measure.
    \label{def:clustering}
\end{definition}

\begin{figure}[!t]
	\centering
	\includegraphics[width=1.02\textwidth]{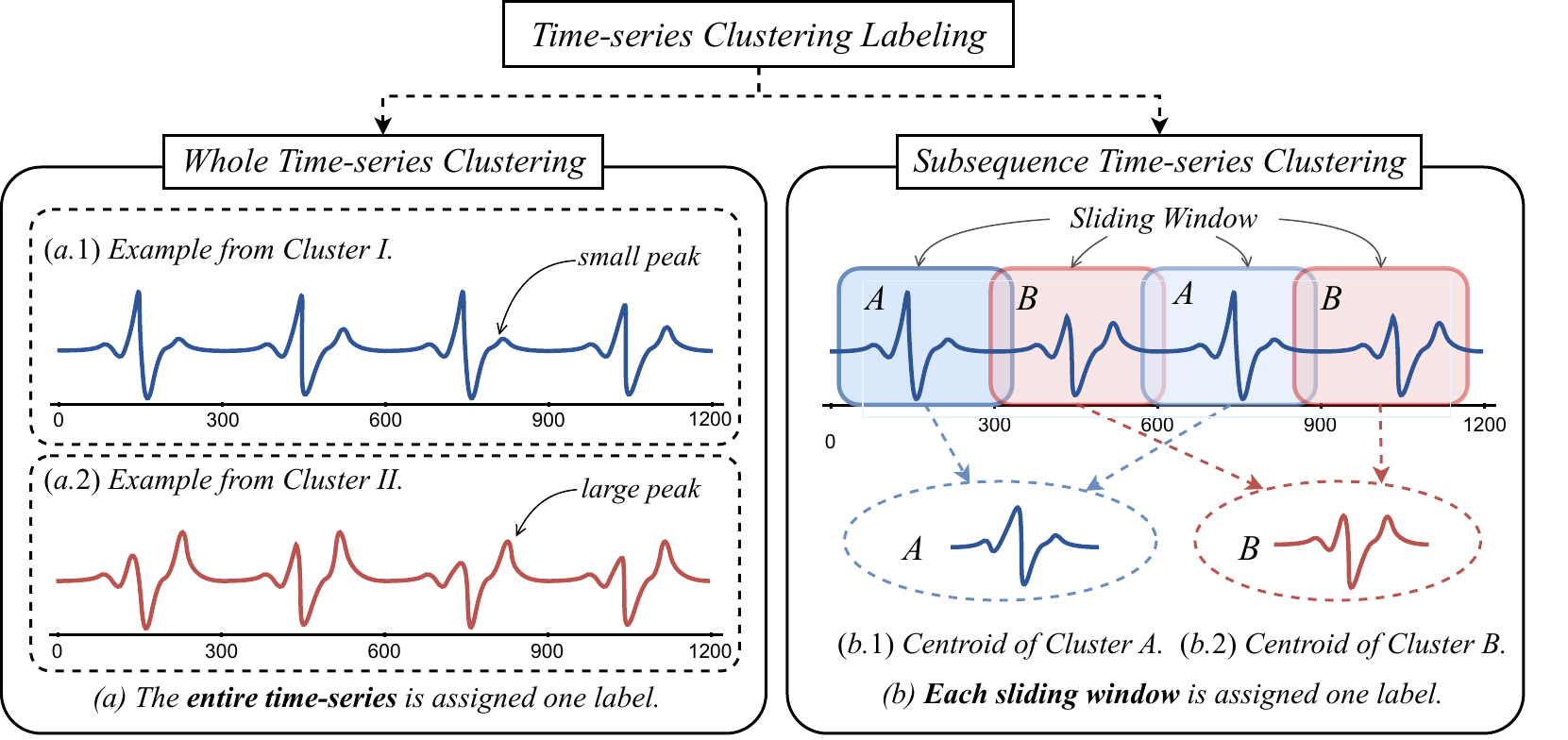}
	\caption{Categories of time-series clustering labeling. Left: whole time-series clustering; right: subsequence time-series clustering.}
	\label{pic:tslabeling}
\end{figure}

Based on the scope, time-series clustering methods can be categorized into 3 main types: \textit{whole time-series clustering}, \textit{subsequence clustering}, \textit{time point clustering} \cite{aghabozorgi2015time}. 

\begin{itemize}
    \item \textbf{Whole Time-series Clustering:} Given a set of time series, individual time series are clustered into distinct groups. In the process, all timesteps will participate in the dissimilarity measure to decide the intra and inter-relationship across instances.
    
    \item \textbf{Subsequence Clustering:} Given a time series, subsequence clustering is defined as the clustering procedure on a set of subsequences obtained by sliding windows.

    \item \textbf{Time Point Clustering:} Given a time series, time point clustering partitions all time points into several groups based on temporal proximity and the similarity in values.
    
\end{itemize}

However, not all of the three types of time-series clustering are meaningful following the discussion in \cite{aghabozorgi2015time}. 
As illustrated in \cite{aghabozorgi2015time}, time point clustering and time-series segmentation exhibit minimal differences. The main difference lies in the fact that some time points may be omitted as noise in the clustering process.
According to the prior studies in \cite{keogh2005clustering}, subsequence clustering may yield random results in the experiments. 
To create meaningful clusters, the process requires specific conditions that are hard to meet in the real world \cite{keogh2005clustering}. The difference between whole time-series clustering and subsequence time-series clustering is shown in Figure \ref{pic:tslabeling}. As a result, we will focus on the \textbf{whole time-series clustering} in this survey.

\begin{figure}[!htp]
	\centering
	\includegraphics[width=\textwidth]{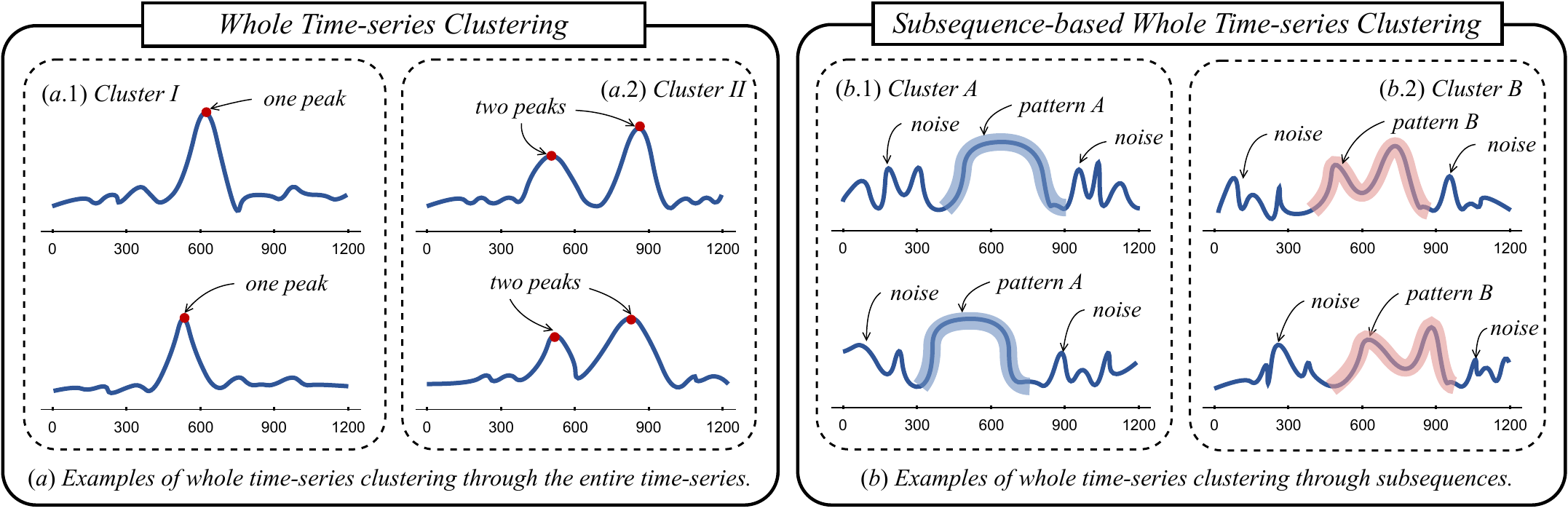}
	\caption{Two main types of whole time-series clustering methods. Left: clustering methods based on entire time series; right: clustering methods based on subsequences.}
	\label{pic:tsclustway}
\end{figure}

We introduce two main types of the whole time-series clustering, identified in Figure \ref{pic:tsclustway} based on the contribution of each time step. Figure \ref{pic:tsclustway}(a) presents the conventional clustering process that treats all time-series steps equally, while Figure \ref{pic:tsclustway}(b) depicts subsequence-based whole time-series clustering which employs a set of subsequences as guidance. Importantly, the latter type differs from what's known as subsequence clustering in that it still produces the clustering labels for the whole time series in the end.

\subsection{Time-series Clustering Pipeline}

Before delving into any specific clustering algorithms, it is important to know the general process of time-series clustering. Building upon the insights presented in prior studies \cite{liao2005clustering, aghabozorgi2015time, alqahtani2021deep}, we summarize the common pipeline of time-series clustering into three parts (shown in Figure \ref{pic:pipeline}): \textit{representation, dissimilarity measure} and \textit{clustering procedure}. This decomposition will become beneficial not only for the comparative evaluation of diverse time-series clustering algorithms, but also help uncover the essence of time-series clustering and develop novel algorithms. 

\subsubsection{Representation}
The \textit{representation} procedure, also named as \textit{data representation} or \textit{feature extraction}, denotes the data format of time series for facilitating downstream tasks. Raw time-series data is the basic format with rich feature information from natural signals, which is widely applied in different clustering algorithms. However, the noise interference from the signal recording poses a challenge in the search of meaningful patterns, and also the large dimension from the original space also significantly increases the time and cost for data analysis. Therefore, an effective approach to extracting meaningful features while retaining the essential information becomes beneficial. The definition of the time-series representation can be found in Definition \ref{def: representation}.

\begin{definition}
Given a time-series data $x_i = \{x_{i1}, x_{i2}, \dots, x_{iT} \}$ with $T$ time steps, we want to find a transformation $ \phi: x_i \rightarrow x^{\prime}_i$, where $x^{\prime}_i = \{x^{\prime}_{i1}, x^{\prime}_{i2}, \dots, x^{\prime}_{iK} \}$ denotes a new representation, specifically $K<T$ in the dimension reduction scenario. The transformation space should retain the essential information from the original space in such a way that, if $x_i$ is similar to $y_i$, then $x'_i$ is similar to $y'_i$, and vice versa.
\label{def: representation}
\end{definition}

\begin{figure}[!t]
	\centering
	\includegraphics[width=\textwidth]{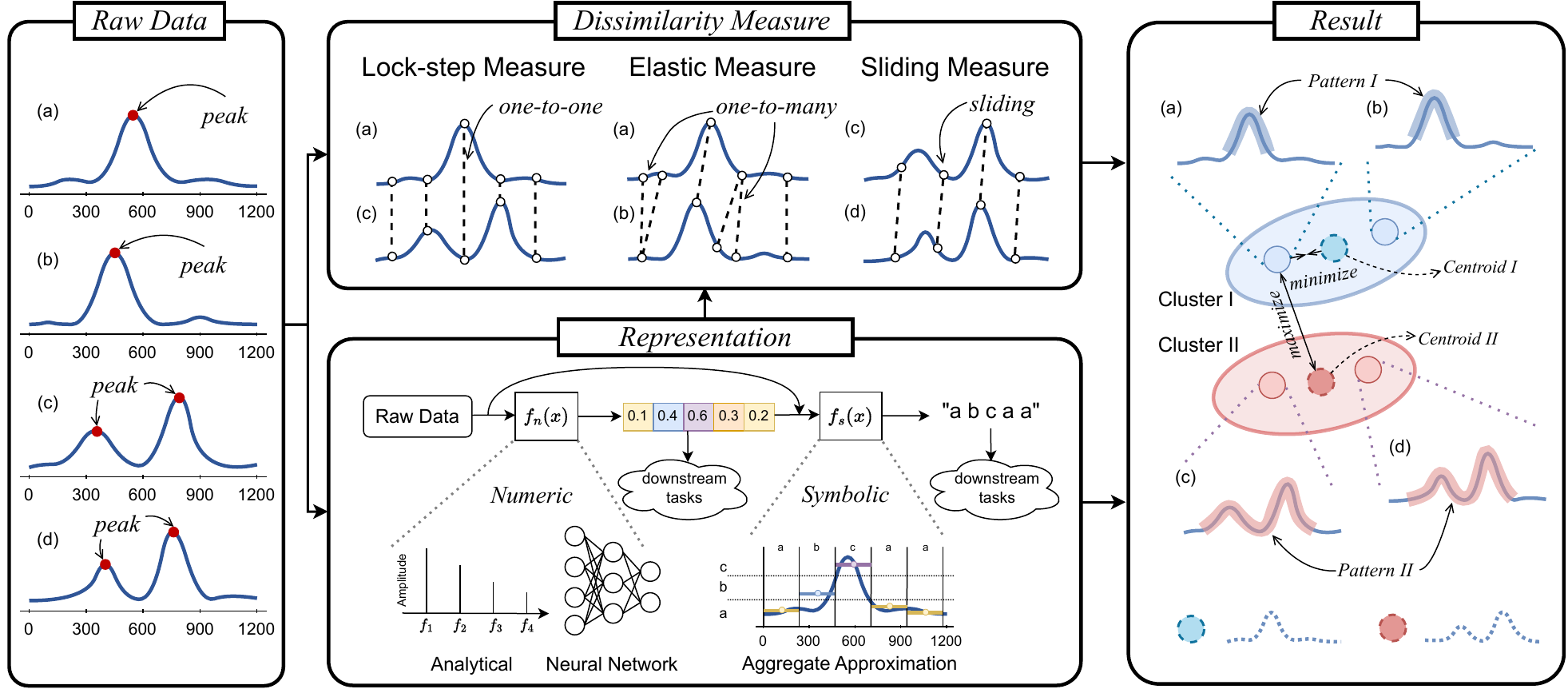}
	\caption{The overview of the time-series clustering pipeline.}
	\label{pic:pipeline}
\end{figure}

As can be seen from Figure \ref{pic:pipeline}, there are two main types of time-series representation: \textit{numeric} and \textit{symbolic}. The numeric time-series representation utilizes real-valued array (univariate) or matrix (multivariate) to denote the feature information in original signals, usually with reduced dimensions. Representative techniques include Discrete Fourier Transform (DFT) \cite{faloutsos1994fast, agrawal1993efficient, kawagoe2002similarity}, Discrete Wavelet Transform (DWT) \cite{chan1999efficient, agrawal1993efficient, kawagoe2002similarity}, Piecewise Linear Approximation (PLA) \cite{keogh1998enhanced}, Piecewise Aggregate Approximation (PAA) \cite{keogh2000simple, yi2000fast}, or neural networks after the emergence of deep learning \cite{guo2017improved,ghasedi2017deep,madiraju2018deep,bo2020structural,zerveas2021transformer,yue2022ts2vec}. Symbolic time-series representation \cite{lin2003symbolic,SFA2012,malinowski20131d}, on the other hand, 
offers the benefit from both dimensionality reduction and the wealth of text-based methodology, e.g. hashing, and sequence matching.
Representative methods are Symbolic Aggregate approXimation (SAX) \cite{keogh2004towards, lin2007experiencing}, indexable Symbolic Aggregate approXimation (iSAX) \cite{lin2007experiencing},  Symbolic Fourier Approximation (SFA) \cite{SFA2012}, and 1d-SAX \cite{malinowski20131d}.
As depicted in Figure \ref{pic:pipeline}, the input of symbolic representation could be either raw time-series data or a transformed representation.

\subsubsection{Dissimilarity Measure}
As shown in Definition \ref{def:clustering}, the primary objective of time-series clustering lies in the process of partitioning the data in a way that time series exhibiting the same pattern should be grouped together. To solve this problem in a mathematical way, the \textit{dissimilarity measure}, also called \textit{distance measure}, is proposed to quantify the proximity or separation relationship between two sequences (could be raw data or transformed representation as shown in Figure \ref{pic:pipeline}). Based on the definition, two identical time series should have a zero dissimilarity measure and a sufficiently large value when they belong to different clusters. With different designs, the dissimilarity measure can capture the similarity information in time, shape, and structure \cite{aghabozorgi2015time}. Generally, there are three major types of dissimilarity measures: lock-step, elastic, and sliding measure. An overview of each is depicted in Figure \ref{pic:pipeline}.

These three types of dissimilarity measures all have their own pros and cons. On the one hand, the one-to-one mapping assumption of the lock-step measure simplifies the comparison between different time series with much less time complexity, i.e., close to $O(n)$, where $n$ is the sequence length. However, on the other hand, dissimilarity measures like Euclidean Distance impose limitations on varying lengths of time-series data, and the one-to-one mapping may suffer from the noise interference inherent in natural signals. On the contrary, the elastic measures provide a flexible alignment in different regions which is robust to various kinds of disturbances and achieves great success in diverse time-series analysis tasks. However, according to the discussion in prior works \cite{paparrizos2020debunking}, a majority of elastic measures do not perform significantly better than sliding measures on the benchmark while there exists a nonnegligible gap between the time consumption. In other words, sliding measures may provide a good trade-off between runtime and accuracy in comparison to lock-step or elastic measures in different cases.

\begin{itemize}
    \item \textit{Lock-step Measure} focuses on one-to-one mapping for two whole time series. The final result is usually calculated by the summation or mean value of errors across each time point. The representative methods are Euclidean Distance (ED) \cite{faloutsos1994fast,giusti2013empirical}, Minkowski (i.e., $L_p$-norm) \cite{batyrshin2013constructing,giusti2013empirical,paparrizos2020debunking}, Lorentzian (i.e., the natural logarithm of $L_1$) \cite{giusti2013empirical,paparrizos2020debunking}, Manhattan \cite{giusti2013empirical,paparrizos2020debunking}, Jaccard \cite{giusti2013empirical,paparrizos2020debunking}. 
    
    \item \textit{Elastic Measure} has been frequently utilized in scenarios when the one-to-one mapping assumption does not hold firmly. Due to the noise interference and the nature of signals, two time-series data samples may be similar but exhibit different distortion in amplitude (scaling) and offset (translation), where the lock-step measures are likely to fail or suffer from performance degradation. To solve this problem, elastic measure methods are proposed to create one-to-many/many-to-one mapping in an ``elastic" way, which can provide a flexible alignment across time points in various regions \cite{paparrizos2020debunking}. Representative elastic measures are dynamic time warping \cite{sakoe1978dynamic,berndt1994using,chu2002iterative,ratanamahatana2005three}, the Longest Common Subsequence (LCSS) \cite{andre1997using,vlachos2002discovering}, Move–split–merge (MSM) \cite{stefan2012move,holder2023review}.
    
    \item \textit{Sliding Measure} is another type of dissimilarity measure, which follows a sliding mechanism to create a global alignment for different sequences. Representative sliding measures include variants of Normalized Cross-Correlation (NCC), such as Shape-based distance (SBD) \cite{paparrizos2015k}. Thanks to the advantage of Fast Fourier Transform (FFT), the cost can be reduced to $O(n\log n)$, which is a significant speed improvement compared to the original version of DTW (time complexity of $O(n^2)$). \cite{paparrizos2020debunking}, 
\end{itemize}

\section{Time-series Clustering Taxonomy}
\begin{figure}[!htp]
	\centering
	\includegraphics[width=\textwidth]{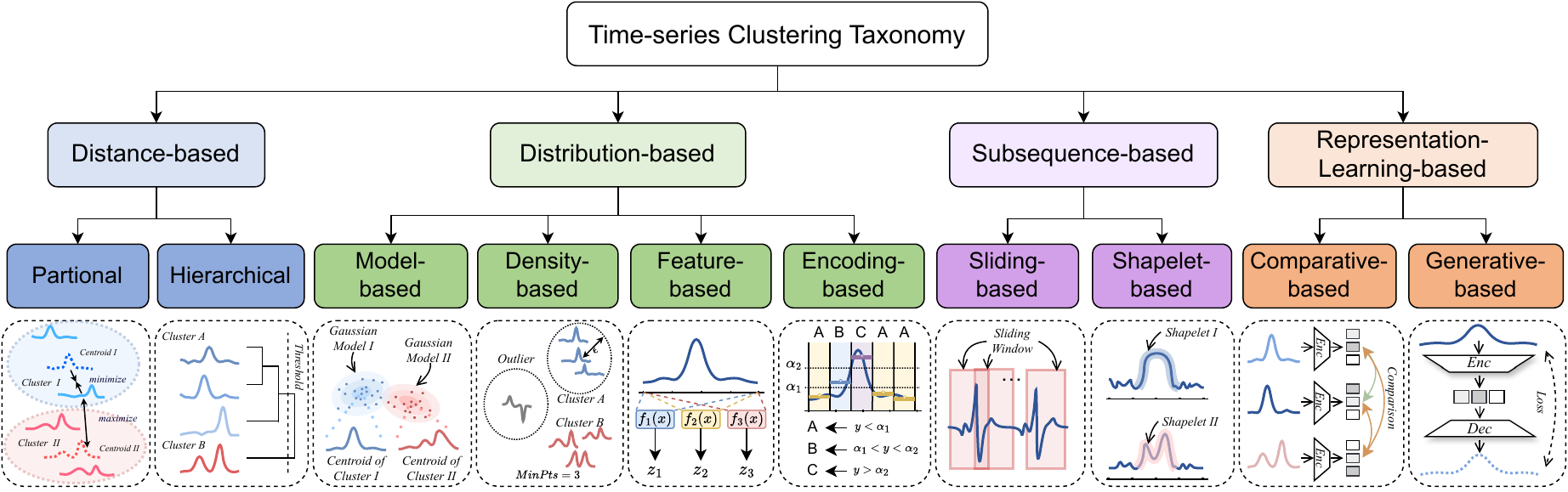}
	\caption{The taxonomy of time-series clustering algorithms.}
	\label{pic:TSClustering_Taxonomy}
\end{figure}

In this section, we describe our proposed taxonomy of the time-series clustering algorithms including both traditional and deep learning-based strategies. All methods are divided into 4 categories: (i) Distance-based, (ii) Distribution-based, (iii) Subsequence-based, and (iv) Representation-learning-based. Figure \ref{pic:TSClustering_Taxonomy} illustrates our proposed taxonomy with a sketch for each category. Next, we review the definition of each category in the following subsections.

\subsection{Distance-based}
As can be seen from the name itself, the crucial concept behind Distance-based methods is the way to measure the distance $Dist$ between two raw time series $X_A$ and $X_B$, where $Dist(X_A, X_B)=0$ if the two time series are the same. Euclidean Distance (ED), as one of the most widely used distance measures in a variety of data formats, has achieved great success in many research fields. However, it is noticed that, for its lock-step design, this efficient distance measure usually suffers from the variance issue inherent in time-series data, e.g., scaling variance, shift variance, occlusion variance, etc. To deal with this problem, many distance measures tailored for time-series data have been proposed, such as Dynamic time warping (DTW) and Shape-based Distance (SBD).

Given the distance between each pair of time series, one can tell that $X_A$ and $X_B$ should be put together if $Dist(X_A, X_B)$ is small and vice versa. To make a proper decision for each time series, there come methods under two second-level categories: partitional and hierarchical models. More detail will be discussed in Section \ref{sec:distance_based}.
\begin{itemize}
    \item The \textbf{partitional} clustering algorithm focuses on partitioning $N$ unlabeled time series into $K$ clusters with centroids. Each time series has the smallest distance from the centroid of the current cluster. The cluster centroids and assignments are usually optimized iteratively through the training process. 

    \item The \textbf{hierarchical} clustering is an approach of grouping objects into clusters through a hierarchy of clusters. Depending on the hierarchy structure. methods can be divided into two types: agglomerative and divisive. In the process, all time series will keep merging (agglomerative) or get separated apart (divisive).
\end{itemize}

\subsection{Distribution-based}
Distribution-based clustering focuses modeling the distribution of the time-series data, which offers guidance in building the boundary between each cluster. It is worth noting that, the distribution here should be considered a general version, including both (i) explicit distribution, e.g., the density of data points; and (ii) implicit distribution, e.g., encoding the time series using a pre-trained dictionary. Here we name four second-level categories: Model-based, Density-based, Feature-based, and Encoding-based models. More detail will be discussed in Section \ref{sec:distribution_based}.

\begin{itemize}
    \item The \textbf{Model-based} clustering methods focus on modeling the explicit distribution of time-series data with learnable parameters, such as the Gaussian Mixture Model (GMM) and Hidden Markov Model (HMM). Each time series can be modeled and represented by a set of parameters, which could serve as guidance for further time-series clustering.
    \item The \textbf{Density-based} clustering methods define clusters as regions of concentrated objects, with cluster boundaries typically occurring in sparse or empty areas. For example, these methods expand clusters in dense neighborhoods and establish boundaries where data points become sparse.
    \item The \textbf{Feature-based} time-series clustering methods are proposed to find descriptive features to represent the characteristics of time series in a global way. Considering the inherent variance issue in time-series data modality, noise interference could pose a challenge in the distance measure between samples. To solve this problem, descriptive features could serve as a noise-robust representation for the clustering purpose.
    \item The \textbf{Encoding-based} time-series clustering methods focus on building a mapping function $\mF: \mX \rightarrow \mZ$ between the original space and the transformed space (also called the latent space), where the latent representation $\mZ$ contains the essential feature information of the original data. When the dimension of the transformed space is smaller than the original space, this process is often viewed as a dimension reduction technique. 
\end{itemize}

\subsection{Subsequence-based}
Subsequence-based clustering is a special case in the whole time-series clustering categories as defined in Section \ref{sec: 2.2}. In the process, representative subsequences will be extracted from the entire time series as an informative pattern for clustering purposes. As the noise perturbation sometimes leads to performance degradation when considering all time steps, by selecting descriptive subsequences models may circumvent this issue in different cases. There are two major second-level categories: Sliding-based and Shapelet-based models, which will be discussed in Section \ref{sec:subseq_based}.

\begin{itemize}
    \item The \textbf{Sliding-based} methods will obtain a set of subsequences using sliding windows. In some cases, these subsequences can be seen as segments of the raw time series to measure the distance from a lower level, while other models focus on multi-stage clustering using subsequence information as a prior.
    \item The \textbf{Shapelet-based} methods, on the other hand, put attention to subsequences that could function as meaningful patterns to represent the raw time series, which are also called shapelets \cite{ye2009time}. Depending on the objective, prior studies either iteratively search for special subsequences or directly learn shapelets from the dataset.
\end{itemize}

\subsection{Representation-learning-based}
Similar to the aforementioned categories like feature-based or encoding-based models, the Representation-learning-based methods also focus on the design of a new representation for the original time-series data. However, they both need explicit mathematical formulas to calculate the numeric value, while the representation-learning-based methods obtain the representation through a learning process. The new representation can serve as an input to some simple clustering models such as k-Means \cite{macqueen1967some}. With the advent of deep neural networks and unsupervised learning strategies, representation-learning-based algorithms have achieved great success in numerous tasks. Depending on the characteristics of learning strategies, we divide three second-level categories: Comparative-based, and Generative-based models. More detail will be discussed in Section \ref{sec:repre_learning_based}.
\begin{itemize}
    \item The \textbf{Comparative-based} time-series clustering methods learn an encoding mapping function $\mathcal{E}: \mX \rightarrow \mZ$ in a comparative way, e.g., comparing similar/dissimilar time-series samples. The encoder mapping function can be learned by a neural network, which is going to be discussed in detail in the following sections.
    \item The \textbf{Generative-based} time-series clustering methods, in contrast to comparative-based clustering methods,  learn the robust representation by casting constraints on the generation output. One good example is the reconstruction task: by jointly learning the encoding mapping function $\mathcal{E}: \mX \rightarrow \mZ$ and the decoding mapping function $\mathcal{D}: \mZ \rightarrow \mX$ in a reconstruction way, deep neural networks are able to find a good latent space for data representation with possible fewer dimensions.
\end{itemize}

\section{Distance-based Methods}
\label{sec:distance_based}
In this section, we explore distance-based clustering methods that leverage and solely depend on measures of dissimilarity, either between data, between clusters, or a combination of both, to group similar elements together. The methods within this category typically do not require specific data representation or feature selection. They operate effectively by utilizing raw time series alone to generate meaningful outputs. Furthermore, this categorization can be further divided into two sub-categories: partitional methods and hierarchical methods. The partitional methods divide a set of $n$ unlabeled time-series data into $k$ clusters and ensure that each cluster comprises at least one time-series data while the hierarchical algorithms group data through establishing a hierarchical structure. In the subsequent sections, we will delve into a more detailed exploration of the concepts and representative clustering methods of these two sub-categories in Section \ref{sec:partition} and \ref{sec:hierarchy}.

\vspace{-0.5em}
\subsection{Partitional Clustering}
\label{sec:partition}
\begin{wrapfigure}{r}{0.5\textwidth} 

    \centering
	\includegraphics[width=\linewidth]{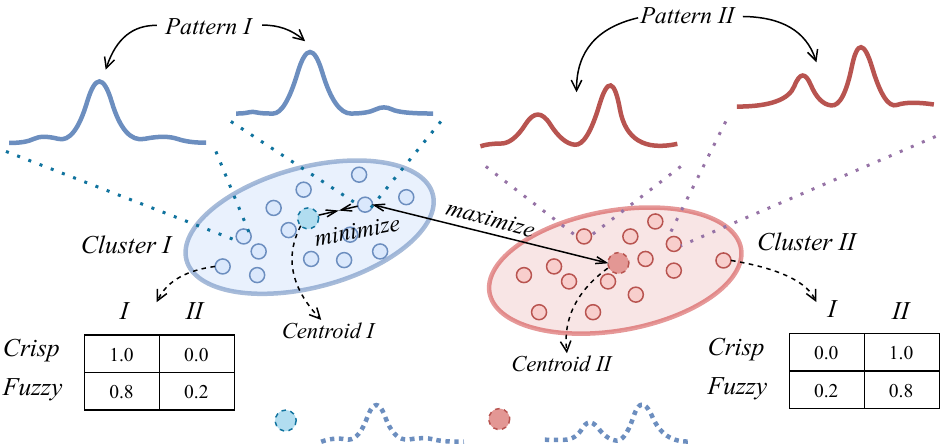}
	\caption{An overview of partitional time-series clustering. Two tables in the figure represent the assignment probability under both crisp and fuzzy partitional methods.}
	\label{pic:partition}
\vspace{-0.5em}
\end{wrapfigure}

The partitional clustering algorithm partitions $n$ unlabeled time-series data into $k$ clusters and guarantees each cluster contains at least one time-series data. Moreover, partitional clustering methods can be classified into two main categories as illustrated in Figure \ref{pic:partition}: crisp partitional methods and fuzzy partitional methods. In crisp partitional methods, each data point is exclusively assigned to a single cluster. In contrast, fuzzy partitional methods associate each data with membership likelihoods and permit each data to belong to multiple clusters simultaneously. In the following sections, we will discuss representative methods belonging to these two categories separately in more detail and related methods can be found in Table~\ref{Partitioning-clustering_table}. \\

\begin{wraptable}{r}{0.45\textwidth}
\footnotesize
\caption{Summary of the Partitional clustering methods.}
\label{Partitioning-clustering_table}
\resizebox{\linewidth}{!}{
\begin{tabular}[t]{llll}
\Xhline{2\arrayrulewidth}  \addlinespace[0.2cm]
Method Name &Distance Measure &Prototype  &Dim\\

\Xhline{2\arrayrulewidth} \addlinespace[0.2cm]
k-Means \cite{macqueen1967some} &ED &k-Means &M \\
k-Medoids \cite{rdusseeun1987clustering} &* &k-Medoids &I \\
k-Medians \cite{jain1988algorithms} &Manhattan &k-Medians &I\\
M-RC \cite{kovsmelj1990cross}         & ED & k-Medoids & M \\
CB-FCM \cite{golay1998new}              & CC-based & FCM &I  \\
NTSA-TC \cite{policker2000nonstationary}  & ED &FCM &I \\
M-k-Medoids \cite{liao2002understanding}     & DTW &k-Medoids &I \\
FSTS \cite{moller2003fuzzy}           & STS distance &FCM &I \\
k-DBA \cite{petitjean2011global}  &DTW & k-DBA  &I \\
K-SC \cite{yang2011patterns}  &STI distance &K-SC &I \\
DFC \cite{ji2013dynamic}             &ED &FCM &I \\
DKM-S \cite{seref2014mathematical}     & Arbitrary & k-Median &M  \\
k-Shape \cite{paparrizos2015k}           & SBD &k-Shape &I \\
k-MS \cite{paparrizos2017fast}  &SBD &k-MS &I \\
m-kAVG \cite{ozer2020discovering} & m-ED &k-Means &M \\
m-kDBA \cite{ozer2020discovering} &m-DTW &k-DBA &M \\
m-kSC \cite{ozer2020discovering} & m-STI &K-SC &M \\
m-kShape \cite{ozer2020discovering} & m-SBD &k-Shape &M\\

\Xhline{2\arrayrulewidth} \addlinespace[0.2cm]
\end{tabular}}
    \begin{tablenotes}
      \scriptsize
      \item  I: Univariate; M: Multivariate; *: Arbitrary;
     \end{tablenotes}

\end{wraptable}

\noindent\textbf{Crisp partitional methods}\\

One of the most widely-used partitional clustering algorithms is k-Means \cite{macqueen1967some}. Initially, given the number of centroids $k$, k-Means randomly selects $k$ time-series data as its initial centroids. Then, k-Means repeatedly uses Euclidean Distance (ED) to compute distances between each object and all centroids, assigns each object to one cluster whose centroid is the closest to that object, and updates each centroid to the mean of objects in that cluster until one of the pre-defined criteria is meet. However, since k-Means is sensitive to the initialization of centroids, \cite{arthur2007k} proposed k-Means++ to improve the performance of k-Means by defining centroids initialization rules which iteratively adding the new centroid based on the probability proportional to objects' distances from their corresponding the nearest previously selected centroids and aims at separating initial centroids from each other as far as possible. 

A variation of k-Means clustering is named k-Medians \cite{jain1988algorithms}. Compared with the iteration described for k-Means, unlike k-Means, instead of using ED as a distance measure and computing the mean for each cluster as its centroid, k-Medians applies Manhattan distance as its distance measure and calculates the median.

Another important clustering algorithm that is worth mentioning is k-Medoids \cite{kaufman2009finding} whose objective is also to minimize the distance between the data point assigned to a cluster and the centroid of that cluster. However, different from k-Means, k-Medoids can use arbitrary distance measures and it uses actual and the most representative data points named medoids as centroids. In other words, the medoid is an actual data point within a cluster, which minimizes the average distance to all other points in that same cluster. Partitioning Around Medoids (PAM) \cite{kaufman2009finding} is one of the most classic and representative methods that belong to the family of the k-Medoids clustering algorithm. Different from k-Means, PAM applies an algorithm named PAM Swap to perform centroid updates. For each medoid $m$, PAM Swap swaps it with a non-medoid point $o$, performs data points assignment, and calculates the total cost of the clustering. If the total cost decreases, the swap is performed and repeated. Otherwise, PAW Swap will undo the last swap and end. Compared with k-Means, incorporating medoids as centroids enhances the algorithm's robustness to the outliers and elevates the interpretability of centroids. Nevertheless, akin to k-Means, the k-Medoids clustering algorithm also requires the number of clusters as its input.

Furthermore, another extended variant of k-Means is called k-DBA \cite{petitjean2011global} which incorporates Dynamic Time Warping (DTW) \cite{sakoe1978dynamic} as distance measure and adapts DTW barycenter averaging (DBA) as its centroid computation. For each refinement, DBA updates each coordinate of the average sequence with the barycenter of its associated coordinates obtained by calculating DTW between the average sequence and each other sequence individually. K-Spectral Centroid (K-SC) \cite{yang2011patterns} clustering algorithm, modified from k-Means, incorporates a scaling and translation invariant (STI) distance measure as well as matrix decomposition technique to update its centroids. 

Another algorithm in this category is k-Shape \cite{paparrizos2015k} and it is a current state-of-the-art time-series clustering algorithm. Different from k-Means, k-Shape utilizes SBD as its distance measure. Adopting normalized cross-correlation and speeding up the computation in the frequency domain, SBD becomes a cheaper and more efficient algorithm compared to some other good-performing algorithms such as DTW. To compute centroids, k-Shape aligns objects within the same cluster towards the cluster’s centroid based on SBD, and, since cross-correlation captures similarity, the objective function becomes finding a new centroid to maximize the sum of squared similarities within a cluster. Then, the optimization problem can be modified into maximization of the Rayleigh Quotient \cite{golub2013matrix} and the new centroid becomes the eigenvector associated with the largest eigenvalue. \\

\noindent\textbf{Fuzzy Partitional methods}\\

Besides partitioning n objects into k clusters in a crisp manner which forces each object to become a part of exactly one cluster, there are fuzzy partitional algorithms that enable one object belonging to more than one cluster to a certain degree. One of the most representative algorithms in this category is named Fuzzy c-Means (FCM) \cite{dunn1973fuzzy} \cite{bezdek2013pattern}. In FCM, each data is assigned a fuzzy membership value for each cluster. These membership values are real numbers, ranging from 0 to 1 and representing the degree of likelihood that the data belongs to a specific cluster. Moreover, for each data $x_i$ and cluster $C_j$ pair, FCM calculates the distance between $x_i$ and the cluster centroid $c_j$ and weights the distance by the membership that $x_i$ belongs to $C_j$. With the goal of minimizing the sum of weighted distances across all data-cluster pairs, FCM iteratively updates membership values and cluster centroids until convergence. After convergence, the clustering configuration is finalized based on ultimate membership values.

\cite{golay1998new} applied FCM to functional MRI and discussed its possible optimizations respecting to three different parameters: 1. data set pre-processing methods; 2. distance measures; 3. cluster numbers. Since they expected that clustering the pixel time courses should be performed based on similarity, they introduced two cross-correlation-based (CC-based) distance and performed comparisons among those two newly proposed measures plus ED to find the best measure.

Considering that many real-world situations involve the problem of short time series (STS) and unevenly sampled time series and motivated by observations in the field of molecular biology, \cite{moller2003fuzzy} proposed a modified fuzzy clustering scheme by applying the proposed STS distance which was able to capture both the similarity of shapes formed by relative change of amplitude and the corresponding temporal information into a standard fuzzy clustering algorithm.

\subsection{Hierarchical Clustering}
\label{sec:hierarchy}
Hierarchical clustering is an approach to grouping objects into clusters by constructing a hierarchy of clusters, which has great visualization power in time-series clustering. Hierarchical clustering methods can be divided into two types: agglomerative and divisive. While agglomerative hierarchical (AH) clustering algorithms start by creating clusters individually for each time series and then iteratively merging small clusters into large clusters until meeting certain criteria, divisive hierarchical (DH) clustering algorithms tend to assign all time series into one cluster and perform division till satisfying certain criteria. Moreover, related methods can be found in Table~\ref{Hierarchical-clustering_table}.

\begin{table}[!htp]
\footnotesize
\caption{Summary of the Hierarchical clustering methods.}
\resizebox{0.65\columnwidth}{!}{
\begin{tabular}[t]{llll}
\Xhline{2\arrayrulewidth}  \addlinespace[0.2cm]
Method Name &Linkage &Distance Measure &Dim\\

\Xhline{2\arrayrulewidth} \addlinespace[0.2cm]
DC-MTS \cite{kakizawa1998discrimination}   &AH + $/$ & KL + Chernoff &M  \\ 
TSC-CBV \cite{van1999cluster}   &AH + $/$ & Root mean square &I \\
local-clustering \cite{qian2001beyond}   &AH + single &Ad hoc distance & I  \\
hError \cite{kumar2002clustering}    &AH + error-adjusted &Gaussian models of errors &I  \\ 
TFDC \cite{shumway2003time}      &AH + $/$ & KL &M  \\
ODAC \cite{rodrigues2008hierarchical} &DH + AH + criteria &RNOMC &M \\
TSC-CN \cite{zhang2011novel}  &AH + average & Triangle similarity + DTW &I \\
TSC-PDDTW \cite{luczak2016hierarchical} &AH + average & DTW + DDTW & I \\
HSM \cite{euan2018hierarchical} &AH + spectral-based &Total variation distance &I \\
TSC-BD \cite{subramaniyan2020generic} &AH + complete &DTW &I \\
TSC-COVID \cite{luo2023time} &AH + Ward's &DTW &I \\

\Xhline{2\arrayrulewidth} \addlinespace[0.2cm]
\end{tabular}}
    \begin{tablenotes}
      \scriptsize
      \centering
      \item  I: Univariate; M: Multivariate; $/$: Non-specify;
     \end{tablenotes}
\label{Hierarchical-clustering_table}
\end{table}%

In AH clustering, since all merging operations are performed among the cluster level and each cluster contains at least one object, an extra similarity or dissimilarity measure should be introduced between two clusters and how to measure and represent the distance between two clusters in AH clustering methods becomes an important topic to explore. Thus, the idea of linkage is used and, in this survey, we list some widely used linkages to measure the distance between clusters \cite{kaufman2009finding}:

\begin{itemize}
  \item \textbf{Single linkage:} In single linkage, given two clusters $C_i$ and $C_j$ of time series, we initially need to compute all pairwise distances $D = \{dist(x, y) \; | \; \forall \: x \in C_i \; and \; \forall \: y \in C_j \}$, where $dist(x, y)$ is a distance function used to measure the distance between two time series $x$ and $y$. Then, the distance between $C_i$ and $C_j$ is defined as the shortest distance in $D$.
  \item \textbf{Complete linkage:} In complete linkage, given two clusters $C_i$ and $C_j$ of time series, we need to calculate all pairwise distances $D$ following the same expression in the single linkage. Then, the distance between $C_i$ and $C_j$ is defined as the longest distance in $D$.
  \item \textbf{Average linkage:} In average linkage, given two clusters $C_i$ and $C_j$ of time series, we should obtain all pairwise distance $D$ following the same expression mentioned above. After that, the distance between $C_i$ and $C_j$ is defined as the average value of $D$.
  \item \textbf{Centroid linkage:} In Centroid linkage, given two clusters $C_i$ and $C_j$ of time series, we initially need to compute the centroid (the mean time series) of each cluster and denote them as $\overline{C_i}$ for cluster $C_i$ and $\overline{C_j}$ for cluster $C_j$. Then, the distance between those two clusters is represented as $dist(\overline{C_i}, \overline{C_j})$.
  \item \textbf{Ward's linkage} \cite{ward1963hierarchical}\textbf{:} In Ward's linkage, considering two clusters $C_i$ and $C_j$, the distance between two clusters, denoted as $\Delta(C_i, C_j)$, is defined as the increase in total within-cluster variance that occurs after merging.
\end{itemize}

It is worth mentioning that although the majority of papers are using AH with different linkages as their clustering methods, \cite{rodrigues2008hierarchical} introduced a method named Online Divisive-Agglomerative Clustering (ODAC) system which applies DH and an agglomerative phase to cluster time-series data streams. ODAC uses DH with Rooted Normalized One-Minus-Correlation (RNOMC) as its dissimilarity measure and a splitting criterion that divides the node based on the most dissimilar pair of streams. Additionally, it incorporates an agglomerative phase to reassemble a previously divided node, responding to variations in the correlation structure between time series.

Compared with partitional clustering algorithms, hierarchical clustering algorithms do not require the pre-definition of the number of clusters, and mentioned by \cite{liao2005clustering}, given the suitable distance measure, hierarchical clustering may cluster time series of varying length. However, hierarchical clustering algorithms are difficult to adjust the clusters after they start and, thus, they are sometimes weak in quality and performance. Moreover, \cite{aghabozorgi2015time} stated that hierarchical clustering has quadratic computational complexity, and thus, due to its computational complexity and poor scalability, it is not optimal to run hierarchical clustering on large datasets.
\section{Distribution-based Methods}
\label{sec:distribution_based}
In this section, we will discuss distribution-based clustering methods, where time-series data are grouped based on their explicit or implicit distribution. Emphasizing on extracting, selecting, learning, and utilizing the distribution of time-series data makes the distribution-based clustering methods be distinguished from the distance-based clustering methods. We further classified the distribution-based clustering methods into four second-level sub-categories: model-based (Section \ref{sec:model-based-section}), density-based (Section \ref{sec:density-based-section}), feature-based (Section \ref{sec:feature-based-section}), and encoding-based (Section \ref{sec:encoding-based-section}) clustering methods. Subsequent sections will provide precise definitions for each sub-category, accompanied by an exposition of representative methods.

\subsection{Model-based}
\label{sec:model-based-section}
Model-based clustering methods focus on modeling the latent distribution of time-series data using sets of parameters. In this way, the distance between two time series can be translated to the comparison between two parameter sets of each. Representative modeling techniques include the Gaussian Mixture Model (GMM), Hidden Markov Model (HMM), Autoregressive Moving Average (ARMA), and Autoregressive Integrated Moving Average (ARIMA). Related methods can be found in Table \ref{model-clustering_table}. \\

\noindent\textbf{Gaussian Mixture Model (GMM)} \\

GMM \cite{bishop2006pattern} is a probabilistic model that approximates the dataset with a mixture of Gaussian distributions. Suppose the number of cluster $K$ is given, the function for a mixture of $K$ multivariate Gaussian distribution is the following:
\begin{align}
    \mathcal{N}(x_i | \mu_j, \Sigma_j) &= \frac{1}{(2\pi)^{D/2} |\Sigma_j|^{1/2}} \exp\left(-\frac{1}{2} (x_i - \mu_j)^T \Sigma_j^{-1} (x_i - \mu_j)\right) \\
    p(x_i) &= \sum_{j=1}^{K} \pi_j \cdot \mathcal{N}(x_i| \mu_j, \Sigma_j)
\end{align}

the multivariate Gaussian density with unknown parameters $(\mu_j, \Sigma_j)$ was denoted as $\mathcal{N}(x_i| \mu_j, \Sigma_j)$ and $D$ represents the dimension of data. For cluster $j$, $\mu_j$ is its mean, $\Sigma_j$ represents its covariance matrix, and $\pi_j$ is its mixture proportion. In order to estimate these parameters, maximizing the complete log-likelihood which has closed-form maxima can be applied. However, since the complete log-likelihood requires the observation of cluster assignments $z_i$ for each $x_i$, which are unknown and should be learned, in the learning stage of GMM, the Expectation-Maximization algorithm (EM) is applied. Starting with a random initialization, we should iteratively execute the Expectation step (E-step) and Maximization step (M-step) until the convergence. In the E-step, the following fuzzy class membership is computed:
\begin{equation}
    \gamma_{ij} = \frac{\pi_j \cdot \mathcal{N}(x_i | \mu_j, \Sigma_j)}{\sum_{k=1}^{K} \pi_k \cdot \mathcal{N}(x_i | \mu_k, \Sigma_k)}
\end{equation}

Meanwhile, in the M-step, the closed-form maxima solutions for parameters are given by the following equations:
\begin{align}
    \pi_j &= \frac{1}{N}\sum_{i=1}^{N} \gamma_{ij} \\
    \mu_j &= \frac{\sum_{i=1}^{N} \gamma_{ij} x_i}{\sum_{i=1}^{N} \gamma_{ij}} \\
    \Sigma_j &= \frac{\sum_{i=1}^{N} \gamma_{ij} (x_i - \mu_j)(x_i - \mu_j)^T}{\sum_{i=1}^{N} \gamma_{ij}}
\end{align}

where $N$ is the number of data. After the completion of model training, for a new data point, we are able to compute the probability of this new data belonging to each cluster based on learned parameters and assign it to a cluster that gives the highest probability. \\

\noindent\textbf{Hidden Markov Model (HMM)} \\

\begin{wrapfigure}{r}{0.42\textwidth} 
\centering
\includegraphics[width=\linewidth]{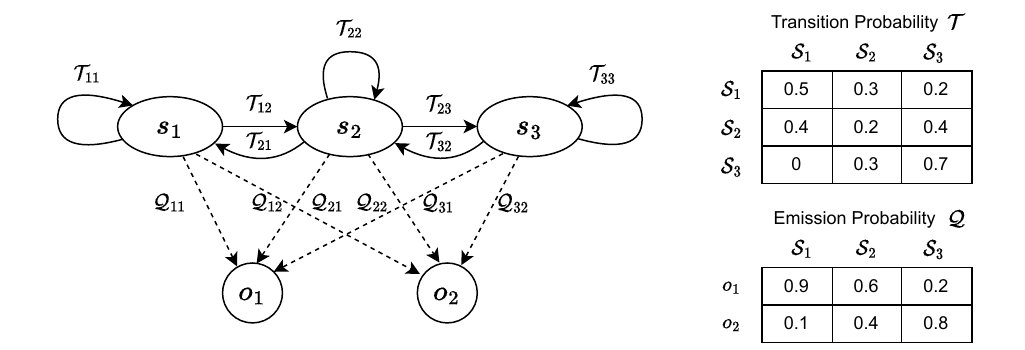}
\caption{An example of Hidden Markov Model (HMM).}
\label{pic:hmm}
\end{wrapfigure}

HMM \cite{bishop2006pattern} is a probabilistic graphical model that aims to infer the underlying hidden states and transitions between states based on the observed data and Figure \ref{pic:hmm} exhibits an example of HMM. There are two primary assumptions behind the HMM. From state space $S = \{s_1, s_2, \cdots, s_N\}$ and observation space $O = \{o_1, o_2, \cdots, o_V\}$, given a sequence of $n$ hidden states written as $X = \{x_1, x_2, \cdots, x_n\}$ and a sequence of $n$ observations denoted as $Y = \{y_1, y_2, \cdots, y_n\}$, the next state $x_{i+1}$ and current observation $y_i$ only depend on current state $x_i$. In HMM, apart from initial probabilities, there are two types of probabilities: transition probabilities and emission probabilities. The transition probability articulates the likelihood of traveling from one hidden state to another while the emission probability is defined as the probability of generating a particular observation given the current hidden state. Additionally, the transition matrix, denoted as $\mathcal{T} \in \mathbb{R}^{N \times N}$, encapsulates the transition probabilities between hidden states while the emission matrix $\mathcal{Q} \in \mathbb{R}^{N \times V}$ captures the emission probabilities from hidden states to observations. To estimate the parameters of HMM, in general, we are trying to find:
\begin{equation}
    \mathcal{T}^*, \mathcal{Q}^* = \operatorname{argmax}_{\mathcal{T}, \mathcal{Q}} P(Y|\mathcal{T}, \mathcal{Q})
\end{equation}

HMM learning also applies the EM algorithm to estimate parameters and one of the most famous instances of the EM algorithm for HMM is called the Baum-Welch algorithm \cite{baum1972inequality}. With trained HMM, for new data, we can infer the most likely sequence of hidden states that generates this observation and perform clustering based on the similarity between the inferred hidden states. \\

\noindent\textbf{Autoregressive Model (AR)} \\

AR \cite{hyndman2018forecasting}, a type of statistical model, believes that a future value in a time series is influenced by its own historical values. This model predicts a value in a time series applying a linear combination of a predefined number of previous values, known as the order of AR, and an AR with order $p$ can be calculated by the following equation:
\begin{equation}
    x_t = \sum_{j=1}^{p} \phi_j x_{t-j} + \varepsilon_t + c
\end{equation}

where, in a time series $x$, $x_t$ denotes the current value at time $t$, $\phi_j$ is the coefficient associated with $x_{t-j}$ which represents the previous value at time $t-j$, $\varepsilon_t$ is the error term at time $t$, and $c$ is a constant term. \\

\noindent\textbf{Moving Average Model (MA)} \\

MA \cite{hyndman2018forecasting}, another classic statistical model, relates the current value in the time series to historical errors. Given the order of MA $q$, MA tends to calculate the value of the time series at time $t$ by incorporating the mean of the time series $\mu$ with a linear combination of previous errors and follows the exact equation as below:
\begin{equation}
    x_t = \sum_{j=1}^{q} \theta_j \varepsilon_{t-j} + \varepsilon_t + \mu 
\end{equation}

where, in a time series $x$, $x_t$ represents the value at time $t$, $\theta_j$ is the parameter associated with $\varepsilon_{t-j}$ which is the error at time $t-j$, and $\varepsilon_t$ denotes the error at time $t$. 

Combining the AR with order $p$ and the MA with order $q$, we can obtain a time series analysis model named Autoregressive Moving Average (ARMA) model \cite{hyndman2018forecasting} which characterizes the current value in a time series based on both previous values and error terms. Extending from ARMA, Autoregressive Integrated Moving Average (ARIMA) \cite{hyndman2018forecasting} integrates differencing with ARMA, empowering it with the capacity to handle non-stationarity in the time-series data.

\begin{table}[!htp]
\footnotesize

\caption{Summary of the Model-based clustering methods.}
\resizebox{0.75\columnwidth}{!}{
\begin{tabular}[t]{lllll}
\Xhline{2\arrayrulewidth}  \addlinespace[0.2cm]
Method Name &Clustering &Model &Distance Measure &Dim\\

\Xhline{2\arrayrulewidth} \addlinespace[0.2cm]
TSC-ARIMA-ED \cite{piccolo1990distance} &AH + complete &ARIMA &ED &I \\
TSC-HISMOOTH \cite{beran1999visualizing} &AH + $/$ & HISMOOTH & $/$ &I \\
TSC-D-HMM \cite{li1999temporal} &4 search levels &HMM &Log-likelihood &M \\
TSC-HMM-DTW \cite{oates1999clustering} &Hybrid &Discrete HMM &DTW &M \\
ICL \cite{biernacki2000assessing} &TSC-ICL &GMM &Log-likelihood &M \\
TSC-AR-HT \cite{maharaj2000cluster} &AH + test-based &AR & Hypothesis test &I \\
MBCD \cite{ramoni2000multivariate} &AH + prob-based &Markov Chain &KL distance &M \\
TSC-LPC-ARIMA \cite{kalpakis2001distance} &PAM &ARIMA &ED &I \\
BHMMC \cite{li2001building} &4 search levels &HMM &BIC &M \\
FCM-SV \cite{tran2002fuzzy} &Modified FCM &GMM &Log-likelihood &I \\
TSC-ARMAM \cite{xiong2002mixtures} &Prob-based  &ARMAs &Log-likelihood &I \\
R-TS-BHC \cite{darkins2013accelerating} &Bayesian HC &Gaussian processes & Dirichlet process & I \\
TSC-HMM-S-KL \cite{ghassempour2014clustering} &PAM &HMM &S-KL divergence &M \\
\Xhline{2\arrayrulewidth} \addlinespace[0.2cm]
\end{tabular}}
    \begin{tablenotes}
      \scriptsize
      \centering
      \item  I: Univariate; M: Multivariate; $/$: Non-specify;
     \end{tablenotes}
\label{model-clustering_table}
\end{table}%

\subsection{Density-based}
\label{sec:density-based-section}

\begin{wraptable}{r}{0.55\textwidth}
\footnotesize
\caption{Summary of the Density-based clustering methods.}
\label{Density-clustering_table}
\resizebox{\linewidth}{!}{
\begin{tabular}[t]{llll}
\Xhline{2\arrayrulewidth}  \addlinespace[0.2cm]
Method Name &Representation &Distance Measure &Dim\\

\Xhline{2\arrayrulewidth} \addlinespace[0.2cm]
DBSCAN \cite{ester1996density}  &Raw &ED & M \\
DENCLUE \cite{hinneburg1998efficient} &Map-oriented &Gaussian kernel &M \\
OPTICS \cite{ankerst1999optics}  &Raw &ED &M \\
FDBSCAN \cite{kriegel2005density}  &Fuzzy object &Fuzzy distance &M \\
DENCLUE 2.0 \cite{hinneburg2007denclue}  &Raw &Gaussian kernel &M \\
D-Stream \cite{chen2007density}  &Raw &Density function &M \\
DPC \cite{rodriguez2014clustering}  &Raw &ED &M \\
TADPole \cite{begum2015accelerating}  &Raw &cDTW &M \\
YADING \cite{ding2015yading}  &Raw &L1 &I \\
ADBSCAN \cite{khan2018adbscan}  &Raw &ED &M \\
\Xhline{2\arrayrulewidth} \addlinespace[0.2cm]
\end{tabular}}
\begin{tablenotes}
  \scriptsize
  \item  I: Univariate; M: Multivariate.
 \end{tablenotes}
\end{wraptable}

In density-based clustering, clusters are subspaces of areas where objects are concentrated, and those dense areas are separated by empty or sparse areas. Diverging from model-based clustering methods, density-based clustering techniques utilize the explicit distribution of time-series data. Notably, many density-based clustering methods can be applied directly to raw time-series data without requiring specific data representations. In the remaining portion of this section, we will discuss some well-known and classic methods within this category and, additionally, more related methods can be found in Table \ref{Density-clustering_table}.

Density Based Spatial Clustering of Applications with Noise (DBSCAN) \cite{ester1996density} expands a cluster if its neighborhood is dense. Given the radius ($\epsilon$) of a circular neighborhood whose center is an object and minimum density threshold (\textit{MinPts}), DBSCAN first separates core points, defined as objects whose circular neighborhoods contain at least \textit{MinPts} objects, from non-core points. Then, DBSCAN forms a cluster by randomly choosing a core point among ungrouped core points and expands that cluster by iteratively adding new core points that are within $\epsilon$ distance from any current core point in that cluster until no core points can be added into that cluster. Moreover, the new cluster includes all non-core points that are within $\epsilon$ distance from any core point in that cluster. Finally, repeat the previous steps besides step one to form the rest of the clusters until every core point belongs to a cluster. The power of DBSCAN is mainly manifested in the following aspects: Firstly, DBSCAN does not require users to determine parameters for the number and shape of clusters; Secondly, DBSCAN is able to handle large datasets; Thirdly, DBSCAN is a robust algorithm that is immune to outliers and noises. 

Ordering Points To Identify the Clustering Structure (OPTICS) \cite{ankerst1999optics}, instead of generating an explicit clustering, forms an augmented ordering of objects and produces a visual representation known as a reachability plot, reflecting the density-based clustering structure of the objects. Extending from DBSCAN, OPTICS introduces extra two terms. The first term is called core-distance and the core-distance of an object $p$ can be calculated through the following equation:
\begin{align}
    \textit{core-dist(p)}= 
\begin{cases}
    \epsilon',& \text{if $p$ is a core point}\\
    \textit{Undefined},              & \text{otherwise}
\end{cases}
\end{align}

where $\epsilon'$ represents the minimum distance demanded to classify $p$ as a core point. Another term is named reachability-distance and the reachability-distance of $p$ with respect to $q$ is defined as:
\begin{align}
    \textit{reachability-dist(p, q)}=
\begin{cases}
    \textit{max(core-dist(q), dist(q, p))}, & \text{if $q$ is a core point}\\
    \textit{Undefined}, &\text{otherwise}
\end{cases}
\end{align}

where $dist(q, p)$ represents the distance between object $q$ and object $p$ and, by default, ED can be used. Compared with DBSCAN, OPTICS provides several advantages such as enabling extracting clusters at different density levels and being less sensitive to the parameters.

DENsity-based CLUstEring (DENCLUE) \cite{hinneburg1998efficient}, a more efficient algorithm compared to DBSCAN, models the influence of each object through a Gaussian influence function with ED, which describes an object's influence within its vicinity. By performing the map-oriented representation, the clustering step of DENCLUE is accelerated by only considering the highly populated cubes and cubes that are connected to them. In order to make the execution more efficient, instead of calculating the overall density of the data space, defined as the summation of all objects' influence functions, DENCLUE computes the local density function which approximates the overall density function by only considering the influence of the neighboring points. Then, the density-attractor, the local maxima of the density function, can be obtained through the hill-climbing procedure. Supported by its firm mathematical basis, besides being invariant to noise, DENCLUE is also efficient on large and high-dimensional data and suitable for finding clusters with various shapes.


Density Peak Clustering (DPC) algorithm \cite{rodriguez2014clustering} is another noteworthy representative in this category and is supported by the idea that the cluster center is distinguished by having a higher density in contrast to its neighbors while maintaining a relatively notable distance from points with higher densities. For each object $i$, DPC calculates the local density $\rho_i$ and the relative distance $\delta_i$ by following the equations below and applying them to plot a decision graph.
\begin{align}
    \rho_i &= \sum_{j} \chi(dist(i, j) - dist_c) \\
    \chi(x) &= 
    \begin{aligned}
    \begin{cases}
        \text{1}, & \text{if $x < 0$}\\
        \text{0}, & \text{otherwise}
    \end{cases}
    \end{aligned} \\
    \delta_i &=
    \begin{aligned}
        \begin{cases}
        \max\limits_j (dist(i, j)), & \text{if $i$ has the highest density} \\
        \min\limits_{j:\rho_j > \rho_i} (dist(i, j)), & \text{otherwise}
        \end{cases}
    \end{aligned}
\end{align}

where $dist(i, j)$ denotes the distance between object $i$ and object $j$ and $dist_c$ represents the cutoff distance. Upon computing $\rho_i$ and $\delta_i$ for each object $i$ in the dataset, DPC constructs a decision graph. This graph utilizes $\rho$ on the x-axis and $\delta$ on the y-axis to determine cluster centers defined as objects with higher $\rho$ and $\delta$ values. Once cluster centers are identified, DPC assigns each remaining point to the same cluster as its nearest neighbor with a higher density. In order to distinguish outliers, for every cluster, DPC identifies a border region, characterized as the collection of objects belonging to that cluster and also lying within a distance of $dist_c$ from objects assigned to other clusters. Within the border region of each cluster, DPC identifies an object $b$ that has the highest local density and records $\rho_b$. By comparing the local density value of each object in the cluster with $\rho_b$, any object with local density that is smaller than $\rho_b$ is considered as an outlier.

Time-series Anytime DP (TADPole) \cite{begum2015accelerating}, a variant of DPC, applies cDTW as its distance measure. Compared to DPC, TADPole requires extra upper bound and lower bound matrices and, as stated by the author, the space complexity is not an issue since the bottleneck of DPC is CPU, and encoding the lower bound matrix into a sparse matrix can reduce the space overhead. During the local density calculation, TADPole prunes the distance computation based on four cases. Meanwhile, it applies a two-phase pruning strategy in calculating the distance to the nearest neighbor with a higher density for each object. As mentioned in the article, the univariate TADPole is able to be extended to the multivariate case with only a few changes.

\subsection{Feature-based}
\label{sec:feature-based-section}
Time series, as one of the basic data formats, represents the changes in signal values across time. However, the noise interference as mentioned above, could pose a challenge in the search for meaningful information especially when the dimension becomes extremely large. To solve this problem, feature-based time-series clustering methods are proposed to find descriptive features to represent the characteristics of time series in a global way. In this section, we will discuss representative methods in detail, and Table \ref{Feature-clustering_table} contains more related methods.

Motivated by the observations that long time series and missing data might cause failures in the many existing clustering algorithms, Characteristic-Based Clustering (CBC) \cite{wang2006characteristic} was proposed. CBC utilizes global structural characteristics measures, combining classical statistical measures with advanced special measures, to cluster time series and they are: trend, seasonality, periodicity, serial correlation, skewness, kurtosis, chaos, non-linearity, and self-similarity. With the assistance of these global representations, CBC is able to significantly decrease the dimensionality of time-series data as well as demonstrate increased robustness against missing or noisy data. As for the clustering methods, for the purpose of obtaining good visualizations, the authors of CBC performed experiments by only applying hierarchical clustering and SOM. Moreover, as stated by the authors, since an appropriate set of features will not only make the computation more efficient but also generate better clustering results, they designed a new technique that is built upon a greedy Forward Search (FS) in order to select the optimized subset of features.

Recognizing the time-consuming aspect of time-series feature engineering, caused by the challenge of navigating through numerous algorithms of signal processing and time-series analysis to extract appropriate and meaningful features from time series, tsfresh \cite{christ2018time}, a widely used and well-known python package, was purposed to accelerate this process. It not only offers 63 time-series characterization methods, which compute a total of 794 time-series features, but also automates the feature extraction and selection based on the FeatuRe Extraction based on Scalable Hypothesis
tests (FRESH) algorithm \cite{christ2016distributed} which demonstrates the ability to scale linearly with the number of features, the quantity of devices/samples, and the number of different time series. However, since the variations in feature calculation costs are affected by their complexities, adjusting the computed features will significantly alter the runtime of tsfresh.

22 CAnonical Time-series CHaracteristics (catch22) \cite{lubba2019catch22} is a widely used feature set distilled from highly comparative time-series analysis (hctsa) \cite{fulcher2017hctsa} toolbox which contains over 7,700 time-series features in its comprehensive version and 4,791 features in the filtered version. Although hctsa has the capacity to select appropriate features for a given application, it is computationally expensive and performs redundant evaluations. Motivated by this observation, the authors of catch22 built a data-driven pipeline that incorporated statistical prefiltering, performance filtering, and redundancy minimization. It is worth mentioning that, in the redundancy minimization phase, they clustered the filtered high-performance features into 22 clusters by applying hierarchical clustering with complete linkage based on the Pearson Correlation (PC) between those features' performance vectors and manually selected features based on their simplicity and interpretability. Executing this pipeline resulted in a canonical set of 22 features, offering a huge improvement in computation efficiency and scalability while only sacrificing 7\% of classification accuracy on average. Moreover, catch22 captures the diverse and representative characteristics inherent in time-series data and its distilled features fall into 7 categories: distribution, simple temporal statistics, linear and nonlinear autocorrelation, successive difference, fluctuation analysis, and others.

\begin{table}[!htp]
\footnotesize
\caption{Summary of the Feature-based clustering methods.}
\resizebox{0.8\columnwidth}{!}{
\begin{tabular}[t]{lllll}
\Xhline{2\arrayrulewidth}  \addlinespace[0.2cm]
Method Name & Clustering &Feature &Distance Measure &Dim\\

\Xhline{2\arrayrulewidth} \addlinespace[0.2cm]
TSC-GC-ED \cite{wang2005dimension} & SOM &Global &ED &I \\ 
CBC \cite{wang2006characteristic} & * & Comprehensive & * & I \\
TSC-SSF \cite{wang2007structure} & * & Statistical & * & M \\
TSC-SF-EU \cite{rasanen2009feature} & k-Means & Statistical & ED & M \\
TSBF \cite{baydogan2013bag} & * & Statistical & ED &M \\
hctsa\cite{fulcher2014highly} & Linear & Comprehensive & * & I\\
FBC \cite{afanasieva2017time} & FBC & Fuzzy & ED &I \\ 
TSA-CF \cite{fulcher2018feature} & * & Comprehensive & * & M\\
tsfresh \cite{christ2018time} & * & Comprehensive & * &M \\
catch22 \cite{lubba2019catch22} & * & Canonical & * &M \\
TSC-CN \cite{bonacina2020time} &Community Detection &Visibility Graph &ED &M \\
TSC-SFLP-ED \cite{choksi2020feature} & k-Means & Statistical + Load Profile & ED &I \\
FeatTS \cite{tiano2021featts,tiano2021feature}&k-Medoid &TSfresh &ED &I \\
TSC-FDDO \cite{zhang2021fault} & Density Peak Search & Comprehensive & ED &I \\ 
TSC-GPF-ED \cite{hu2021classification} & k-Means & Global + Peak & ED &I \\
AngClust \cite{li2022angclust} & Affinity propagation & Angular & PC &M \\
FGHC-SOME \cite{wunsch2022feature} & SOM & Statistical & PC &M \\
FTSCP \cite{enes2023pipeline} & * & Comprehensive & ED &M \\ 
TSC-VF \cite{wu2023imaging} & * & Visual & * & I \\

\addlinespace[0.2cm]
\Xhline{2\arrayrulewidth} \addlinespace[0.2cm]
\end{tabular}}
    \begin{tablenotes}
      \scriptsize
      \centering
      \item  I: Univariate; M: Multivariate; *: Arbitrary;
     \end{tablenotes}
\label{Feature-clustering_table}
\end{table}%

\subsection{Encoding-based}
\label{sec:encoding-based-section}

\begin{figure}[!htp]
	\centering
	\includegraphics[width=0.8\textwidth]{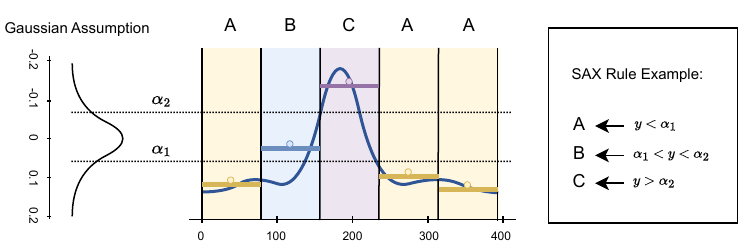}
	\caption{An example of Symbolic Aggregate approXimation (SAX).}
	\label{pic:sax}
\end{figure}

The encoding-based time-series clustering methods focus on building a mapping function $\mF: \mX \rightarrow \mX^{\prime}$ between the raw data space and the transformed space (also called the latent space). This process could be viewed as an \textit{encoding} process.  Compared with time-series clustering methods using raw data, the new representation $\mX^{\prime}$ after encoding captures the crucial information from the original signals with possibly much fewer dimensions. With this advantage, the new representation $\mX^{\prime}$ usually becomes a better option for the downstream procedure with better performance and less computation cost. We summarize all the encoding-based methods in Table \ref{Encoding-clustering_table}.

It is noteworthy that although methods from both categories generate a new sequence of values for clustering purposes, there is one crucial difference between the definition of feature-based clustering and encoding-based clustering: feature-based clustering methods put attention on manually selecting descriptive features based on human knowledge of the current field, e.g., activation strength and delay in fMRI data, while encoding-based methods will use explicit functions to automatically decompose the signal components such as Discrete Fourier Transform, and Piecewise Linear Approximation (PLA).

Generally, the mapping function $\mF$, which is the crucial part of the encoding-based time-series clustering methods, could be represented by a predefined mathematical formula, e.g., Fourier Transform. Representative methods will be discussed in the following paragraphs.


\begin{table}[!htp]
\footnotesize
\caption{Summary of the Encoding-based clustering methods.}
\resizebox{0.8\columnwidth}{!}{
\begin{tabular}[t]{lllll}
\Xhline{2\arrayrulewidth}  \addlinespace[0.2cm]
Method Name &Clustering &Encoding &Distance Measure &Dim\\

\Xhline{2\arrayrulewidth} \addlinespace[0.2cm]
MKM \cite{wilpon1985modified} &Modified k-Means &LPC coefficients &Modified Itakura distance &I \\
CA-CTS \cite{shaw1992using} &AH &PCA &ED & M \\
CDM \cite{keogh2004towards} &Hierarchical &SAX & CDM &I \\
I-kMeans \cite{lin2004iterative} &k-Means / EM &Wavelets &ED &I \\
CTS-CD \cite{bagnall2005clustering} &k-Means / k-Medoids &Clipped &ED &I \\
TSC-CR-LB \cite{ratanamahatana2005novel} &k-Means &Clipped & LB\_clipped &I \\
SAX \cite{lin2007experiencing} & Hierarchical / Partitional &SAX &MINDIST & I \\
TSC-ICA-SDA \cite{guo2008time} &Modified k-Means &ICA &Unknown &I \\
ICTS-FC \cite{aghabozorgi2012incremental} &FCM & DWT &LCS &I \\
TSC-DSA-DTW \cite{gullo2012time} &k-Means &DSA &DTW &I \\
TTC \cite{aghabozorgi2014hybrid} &Hybrid &PAA &ED + DTW &I \\
SAX Navigator \cite{ruta2019sax} &AH &SAX &MINDIST variant &I \\ 
SPIRAL \cite{lei2019similarity} & DTW-preserving   & SPIRAL  & DTW  & M \\

\Xhline{2\arrayrulewidth} \addlinespace[0.2cm]
\end{tabular}}
\label{Encoding-clustering_table}
\vspace{-1em}
\end{table}%


\vspace{1em}
\noindent\textbf{Discrete Fourier Transform (DFT)}

Discrete Fourier Transform (DFT) is one of the most popular mathematical techniques in digital signal processing (DSP), which is also the main focus of our time-series field. Generally, Discrete Fourier Transform converts a sequence of N raw data $\mathbf{x}=\{x_0, x_1, \cdots, x_{N-1}\}$, into a new sequence of complex numbers $\mathbf{X}=\{X_0, X_1, \cdots, X_{N-1}\}$ in the frequency domain as shown in Eq. \ref{DFT}. It is found that the new representation from the frequency domain has a very nice interpretability: low-frequency components usually capture the signals in the raw data which changes slowly over time, while high-frequency components would put more attention on the rapid changes. In real cases, low-frequency components have a high chance of revealing the dominant trends in the raw data and the background noise could be represented by the high frequency. 

\begin{equation}
    X_k = \sum_{n=0}^{N-1}x_n \cdot \mathrm{e}^{-\frac{i2\pi}{N}kn}
    \label{DFT}
\end{equation}

\vspace{1em}




\noindent\textbf{Principal Component Analysis (PCA)}

PCA \cite{pearson1901liii} is a classic and well-known technique for dimensionality reduction. The primary objective of PCA is to encode a high-dimensional dataset into a lower-dimensional representation while maintaining as much original information as possible. Prior to performing PCA, it is crucial to make the mean of each variable become 0 since this pre-processing step will eliminate bias in principal components. After obtaining these mean-adjusted data, a symmetric covariance matrix $A$ is calculated in order to explore any correlation between variables. On top of that, PCA requires the calculation of eigenvectors $\mathbf{v}$ and eigenvalues $\lambda$ (Equation $~\eqref{eq:eigenvector}$) for the covariance matrix $A$ to determine the principal components which are orthogonal and capture the information about directions in the original data where variation occurs. It is worth mentioning that the first principal component encapsulates the direction of the highest variation in the data, the second principal component represents the direction of the second highest variation, and so forth. Since the definition of eigenvectors of the covariance matrix is the same as principal components, PCA sorts the eigenvectors in descending order based on their corresponding eigenvalues, forms a feature matrix $U$ with the eigenvectors in the columns by deciding which eigenvectors to keep, and projects the matrix $X$ with the mean-adjusted original data in rows onto the selected principal components to obtain the final data $X'$ based on Equation $~\eqref{eq:eigenvector}$.
\begin{align}
    A\mathbf{v} &= \lambda\mathbf{v}, \quad
    X' = X \cdot U \label{eq:eigenvector}
\end{align}

\noindent\textbf{Piecewise Aggregate Approximation (PAA)}

Given a time series with length $n$, denoted as $x = \{x_1, x_2, \ldots, x_n\}$, PAA \cite{keogh2001dimensionality} encodes it into another vector $p = \{p_1, p_2, \ldots, p_w\}$ where $w$ is a user-specified input and $w \ll n$. PAA separates the time series into $w$ equal-sized frames and calculates the mean for each frame. The vector consists of these computed mean values, serving as the PAA representation (see Eq. \ref{eq: paa}).
\begin{equation}
    p_i = \frac{w}{n} \sum_{j=\frac{n}{w} (i-1) + 1}^{\frac{n}{w} i} x_j,
    \label{eq: paa}
\end{equation}
where each element $p_i \in p$ can be calculated by the above equation based on $x_j \in x$.

\vspace{1em}

\noindent\textbf{Symbolic Aggregate Approximation (SAX)}

SAX \cite{lin2003symbolic} is another noteworthy representation method that encodes the time series in a symbolic manner and Figure \ref{pic:sax} exhibits an example of SAX. Given a time series with length $n$, SAX will reduce it into a string of length $w$, where $w$ is defined by users and $w \ll n$. After performing Z-normalization on the time series, SAX applies PAA representation which splits the time series into $w$ equal-sized frames, computes the mean for each frame, and concatenates these values into a vector $P = \{p_1, p_2, \ldots, p_w\}$. Since the normalized time series exhibits the strongly Gaussian distribution, given the alphabet size $\alpha$, SAX determines the breakpoints, denoted as a sorted list $B = \{b_1, b_2, \ldots, b_{\alpha-1}\}$ ($b_0$ and $b_\alpha$ are defined as $-\infty$ and $\infty$ correspondingly), which separate the Gaussian curve $\mathcal{N}(0, 1)$ into $\alpha$ equal-sized regions and each region under the curve is equal to $\frac{1}{\alpha}$. Based on the obtained PAA representation and breakpoints, SAX contrasts all PAA coefficients to breakpoints in order to find the specific range that each PAA coefficient falls into and assigns the range-associated alphabet to represent that PAA frame. The final representation is denoted as $S = \{s_1, s_2, \ldots, s_w\}$ and each $s_i$ is calculate by the following function:

\begin{equation}
    s_i = alpha_j, \; iif \; b_{j-1} \leq p_i < b_j
\end{equation}

where $alpha_j$ represents the alphabet associated to range from $b_{j-1}$ to $b_j$.
\section{Subsequence-based Methods}
\label{sec:subseq_based}
Subsequence-based clustering is a special case in the whole time-series clustering categories as mentioned in Section \ref{sec: 2.2}. Unlike previous clustering methods such as partition-based or encoding-based, subsequence-based methods apply one or several subsequences to represent the entire time series, e.g., time series with similar shapelets can be clustered into the same group. It is important to note that, contrary to the subsequence clustering in Section \ref{sec: 2.2}, subsequence-based clustering methods still aim to assign a single label for an entire time series, which exactly aligns with whole time-series clustering in our definition. The subsequence-based methods have two second-level categories:  sliding-based (Section \ref{sec:sliding_based}) and shapelet-based (Section \ref{sec:shapelet_based}) algorithms. Related methods can be found in Table \ref{tab: Subsequence-clustering_table}.

\begin{table}[!htp]
\footnotesize
\caption{Summary of the Subsequence-based clustering methods.}
\resizebox{0.6\columnwidth}{!}{
\begin{tabular}[t]{llll}
\Xhline{2\arrayrulewidth}  \addlinespace[0.2cm]
Method Name & Second Level & Distance Measure &Dim\\

\Xhline{2\arrayrulewidth} \addlinespace[0.2cm]
U-Shapelets \cite{zakaria2012clustering} & Shapelets & LN-ED &I \\ 
SUSh \cite{ulanova2015scalable} & Shapelets & ED & I\\
USLM \cite{zhang2016unsupervised}  & Shapelets & Soft minimum function & I \\
LDPS \cite{lods2017learning}  & Shapelets & Convolutional score & I \\
USSL \cite{zhang2018salient} & Shapelets & ED & I  \\
FOTS \cite{fotso2020frobenius} & Shapelets & FOTS & I \\
LSH-us \cite{luo2020accelerated} & Shapelets & LN-ED & I \\
Trendlets \cite{johnpaul2020trendlets} & Shapelets & Ward Distance & I \\
ShapeNet \cite{Li2021shapenet} & Shapelets & ED & M \\
SPF \cite{li2021time} & Shapelets & Boolean & I \\
MUSLA \cite{zhang2022multiview} & Shapelets & ED & M  \\
TSC-BLU \cite{kafeza2022time} & Shapelets & LN-ED & M \\
CDPS \cite{el2023constrained} & Shapelets & DTW + ED & I \\
CSL \cite{liang2023contrastive} & Shapelets & Multiple Distance & M \\

\addlinespace[0.1cm]\hline \addlinespace[0.1cm]
MSCPF \cite{ren2017sliding}  & Sliding Window & ED & M \\
TS3C \cite{guijo2020time} & Sliding Window & ED & I  \\
TCMS \cite{li2022time}  & Sliding Window & ED & I \\
MPNCMI \cite{li2023time} & Sliding Window & ED & I \\

\addlinespace[0.1cm]

\Xhline{2\arrayrulewidth} \addlinespace[0.2cm]
\end{tabular}}
\label{tab: Subsequence-clustering_table}
\end{table}%

\subsection{Sliding-based}
\label{sec:sliding_based}
Contrary to previous methods which attempt to compare time series across all time steps, sliding-based methods tackle this problem in a smaller scope each time: sliding windows. As illustrated in Section \ref{sec: 2.1}, sliding windows are a set of subsequences extracted by sliding a ``window" in the same length. These subsequences could then be applied to provide feature information or directly function as one way of time-series similarity measure. There are two major directions under this category: (i) Matrix Profile and (ii) Subsequence Clustering. Related methods are summarized in Table \ref{tab: Subsequence-clustering_table}.

\vspace{1em}
\noindent\textbf{Matrix Profile}

Matrix profile \cite{yeh2016matrixi,zhu2016matrixii} is a scalable algorithm for time series all-pairs-similarity-search in the subsequence level. Using a sliding window mechanism, the algorithm could extract possible subsequences and make matches efficiently, which is helpful for both motif discovery and solving discord problems. There are two primary components: matrix profile and matrix profile index. The definition is shown below.
\begin{definition} [Matrix Profile \cite{yeh2016matrixi}]
\label{def: MatrixProfile} 
A matrix profile $P_{AB}$ of time series $A$ and $B$ is a vector of the normalized Euclidean distances between each pair in the similarity join set $J_{AB}$, i.e., a set contains all subsequences pairs $(A_i, B_j)$ obtained from sliding windows where each pair in the set is the nearest neighbor.
\end{definition}

\begin{definition} [Matrix Profile Index \cite{yeh2016matrixi}]
\label{def: MatrixProfileIndex} 
A matrix profile index $I_{AB}$ of time series $A$ and $B$ is a vector of indexes where $I_{AB}[i]=j$ if the pair $(A_i, B_j)$ in the set is the nearest neighbor.
\end{definition}

Given the definition, the matrix profile $P_{AB}$ and the profile index $I_{AB}$ can be seen as metadata or a distance measure of special format between two time series. TCMS \cite{li2022time} defines the degree of correlation between two time series as the number of matched subsequences, i.e., the most similar subsequences found by Matrix Profile. Then the time-series dataset can be transformed to a graph network where each vertex is a time series and the edge is represented by the correlation. To solve the clustering problem, community detection is adopted for partitioning the network. MPNCMI \cite{li2023time} adopts the Matrix Profile in the similarity measure process to find similar subsequences. The normal cloud model is applied to the subsequence pairs to filter the pairwise information. Community discovery is conducted on the complex network to acquire the clustering results.

\vspace{1em}
\noindent\textbf{Subsequence Clustering}

Prior studies also explore the possibilities of clustering time series through multi-stages. Given the subsequences from sliding windows, subsequence clustering can be first conducted to obtain preliminary feature information before the whole time-series clustering in the final stage. MSCPF \cite{ren2017sliding} proposes a sliding window-based multi-stage clustering algorithm using dynamic sliding time windows (DSTW). Subsequence clustering is performed in the first stage for segment information. In the second stage, the small segmented clusters are aggregated to obtain the final clustering results. This multi-stage clustering also provides a foundation for further time-series forecasting tasks. TS3C \cite{guijo2020time} adopts a similar idea. In stead of the previous subsequence clustering strategy after sliding windows, a least-squares polynomial segmentation strategy is adopted. Then each time series can be mapped through the feature information of all segments for further clustering.

\subsection{Shapelet-based}
\label{sec:shapelet_based}
\begin{figure}[!htp]
	\centering
	\includegraphics[width=0.9\textwidth]{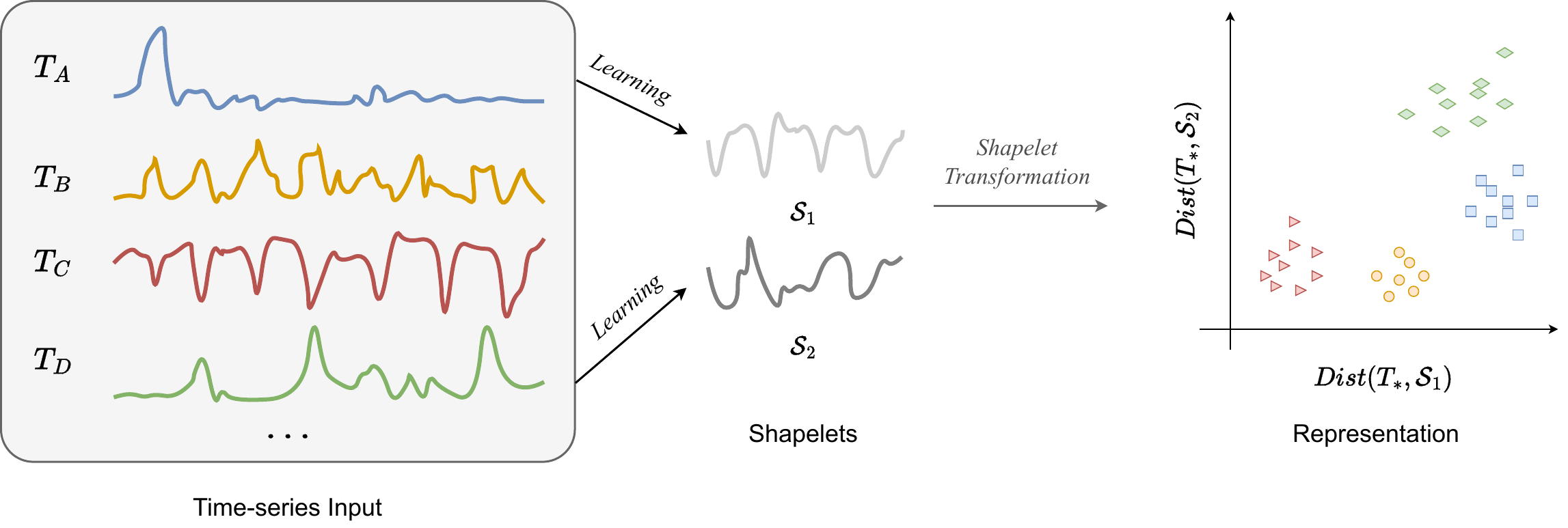}
	\caption{The overview of a shapelet transform pipeline \cite{Li2021shapenet}.}
	\label{pic:shapelet}
\end{figure}

The shapelet-based method is another category of time-series clustering in the subsequence level. 
It is found that some repeated subsequences could be exploited as meaningful patterns for time-series representation, which is called \textit{shapelet} \cite{ye2009time}. With the development of data mining techniques, more and more shapelet-based methods have emerged to search for robust shapelets in time-series clustering tasks \cite{zakaria2012clustering,lods2017learning,zhang2016unsupervised,ulanova2015scalable,fotso2020frobenius,zhang2018salient}. In this section, we follow the definition of shapelet from \cite{ye2009time, zakaria2012clustering}. Related methods can be found in Table \ref{tab: Subsequence-clustering_table}. In Figure \ref{pic:shapelet}, we show an overview of the general shapelet transform pipeline. With a set of learned shapelets, each time series can be projected to a new representation space by measuring the distance between the entire sequence and the shapelets. Time series from the same group might end up being closer in this new space, serving as critical guidance for the clustering process.

\begin{definition}[Shapelet] 
\label{def: shapelet}
Given a time-series dataset that consists of multiple classes, a set of shapelets $\Acute{S}$ are subsequence time series that are highly predictive of the time-series classes.
\end{definition}

U-Shapelets \cite{zakaria2012clustering} proposes a novel method for shapelet searching from the unlabeled time-series dataset, which tries to maximize the separation gap between clusters using different subsequences. A distance map can be easily obtained by calculating the distance between the subsequence distance $sdist$ between the shapelet and the given time series. Such distance maps can be further applied to different clustering algorithms like K-Means. However, the overall time complexity makes U-Shapelets search intractable and not applicable on large datasets. To solve this problem, Scalable U-Shapelet (SUSh) \cite{ulanova2015scalable} is proposed to allow scalable shapelet discovery without significant performance loss. SAX representation with random masking is applied to speed up the similarity measure between time series, where the collision number in comparison can be viewed as an important predictor of qualities. Experiments have demonstrated that time complexity can be reduced by two orders of magnitude. 

Unsupervised Shapelet Learning Model (USLM) \cite{zhang2016unsupervised}, on the contrary, designs an efficient shapelet learning strategy for unlabeled time-series data, instead of the time-costly searching as before. In the process, the candidate shapelet $\Acute{S}$, classification boundary $W$ and pseudo-class label $Y$ can be jointly optimized through a least square minimization problem with shapelet transform. Learning DTW-Preserving Shapelets (LDPS) \cite{lods2017learning} proposes a novel learning strategy for DTW-preserving shapelets without labels. The overall objective is to approximate the DTW in the original space using the shapelet transformation in Euclidean space. It is noteworthy that the concept of shapelet transform and shapelet match share a similar form compared with the Convolutional Neural Network, which naturally brings the convolutional variant of LDPS for application.

With the advent of deep neural networks, more and more studies focus on exploring the possibility of deep learning strategies. CSL \cite{liang2023contrastive} proposes a multi-grained contrastive strategy for shapelet learning on multivariate time-series data. In the framework, the shapelet is represented as learnable parameters and a shapelet transformer is adopted as the encoder to acquire the latent embedding. Considering the issue of variable length, multi-scale alignments are designed to preserve consistency across each level. As an unsupervised representation learning algorithm, CSL could produce a shapelet-based representation suitable for various time-series downstream tasks.
\section{Representation-learning-based Methods}
\label{sec:repre_learning_based}
In recent decades, the rise of deep learning has introduced numerous time-series clustering methods, leveraging the powerful representation capabilities of deep neural networks. In this section, we will first outline the key components of representation-learning-based methods (Section \ref{sec:represent_learning_overview}), and then review individual representative algorithms, consisting of two sub-level categories: Comparative-based (Section \ref{sec:comparative_based}) and Generative-based (Section \ref{sec:generative_based}).  A comprehensive summary of the surveyed methods is provided in Table \ref{tab: representation-learning}.

\begin{figure}[!htp]
\centering
\begin{subfigure}[t]{0.48\linewidth}
\centering
\includegraphics[width=\linewidth]{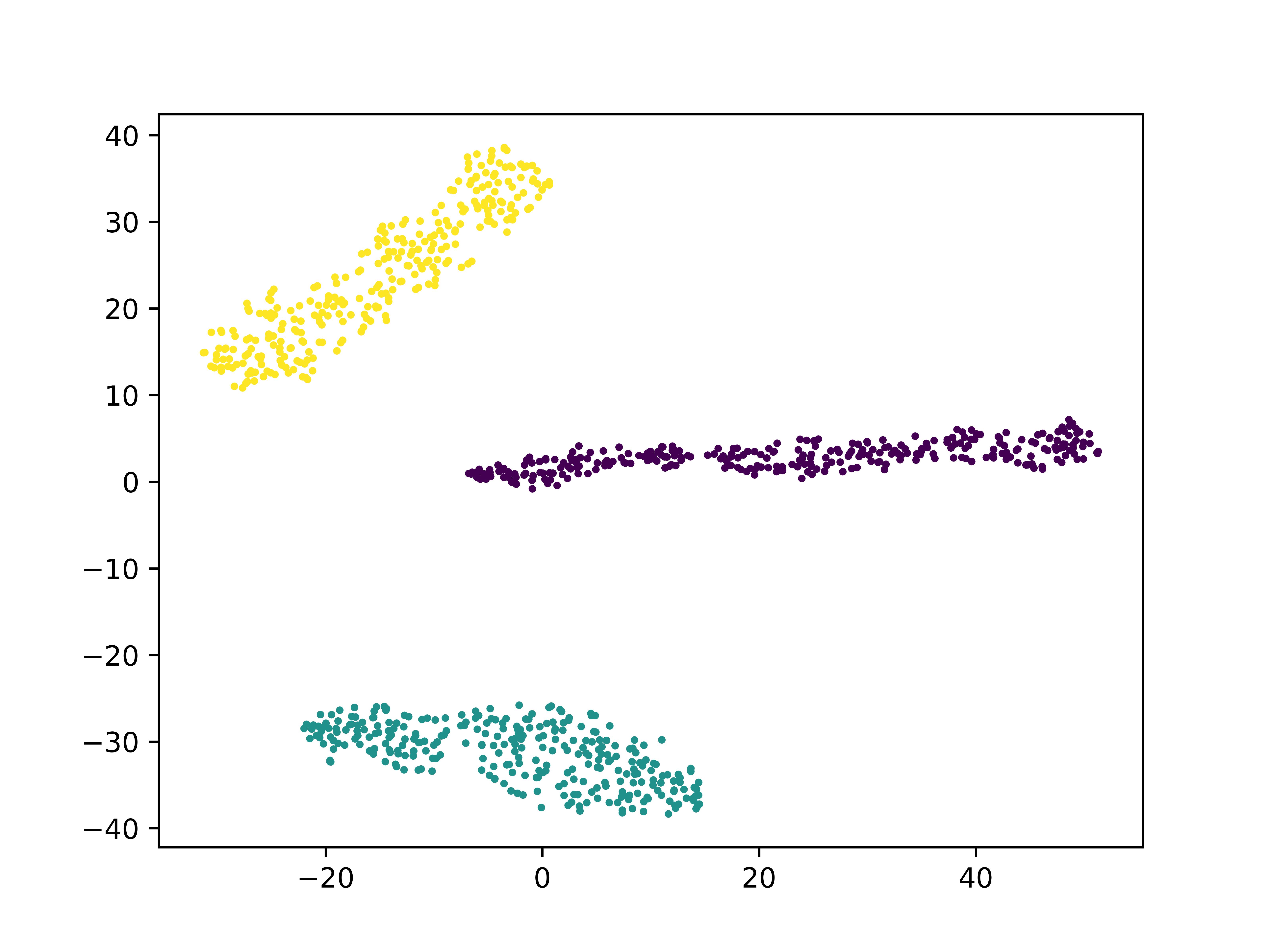}
\caption{CBF Dataset}
\end{subfigure}
\begin{subfigure}[t]{0.48\linewidth}
\centering
\includegraphics[width=\linewidth]{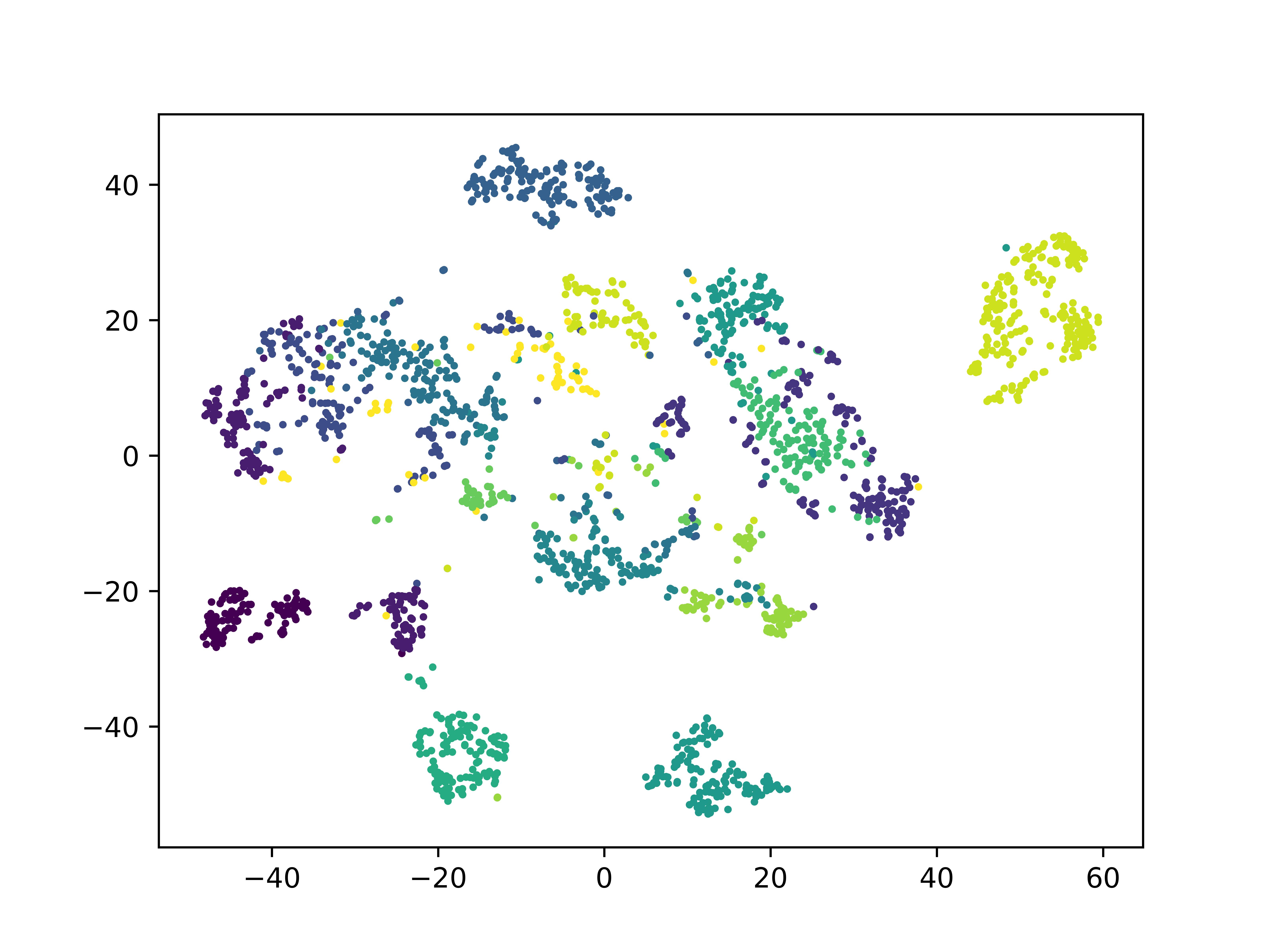}
\caption{FacesUCR Dataset}
\end{subfigure}
\caption{The T-SNE visualization of the encoded representation (learned by representation-learning-based strategy) on the UCR datasets. Left: CBF Dataset with 3 clusters. Right: FacesUCR Dataset with 14 clusters.}
	\label{pic:tsne}
\end{figure}

\begin{table}[!htp]
\footnotesize

\caption{Summary of the Representation-learning-based clustering methods.}

\begin{tabular}[t]{lcccccc}
\Xhline{2\arrayrulewidth}  \addlinespace[0.2cm]
 & Second Level & Strategy & Backbone & Label & Dim\\

\Xhline{2\arrayrulewidth} \addlinespace[0.2cm]

DEC \cite{xie2016unsupervised} & Comparative-based & CLS & AE    & $U^*$ & $I^*$ \\ 
ClusterGAN \cite{ghasedi2019balanced} & Comparative-based & ADV & GAN   & U & $M^*$ \\
TCGAN \cite{huang2023tcgan} & Comparative-based & ADV & GAN   & U & M \\
T-Loss \cite{franceschi2019unsupervised}   & Comparative-based & CNRV  & CNN   & U & M \\
TS2Vec \cite{yue2022ts2vec}         & Comparative-based  & CNRV    & CNN   & U & M \\
SleepPriorCL \cite{zhang2021sleeppriorcl}  & Comparative-based  & CNRV   & CNN   & U & M \\
TF-C \cite{zhang2022self}                & Comparative-based   & CNRV    & CNN   & U & M \\
BTSF \cite{yang2022unsupervised}         & Comparative-based   & CNRV    & CNN   & U & M \\
MHCCL \cite{meng2023mhccl}               & Comparative-based   & CNRV    & CNN   & U & M \\
Ts2DEC \cite{ienco2019deep}              & Comparative-based   & CNRV    & CNN   & $Se^*$ & I \\
DCRLS \cite{xiao2023deep}                & Comparative-based   & CNRV    & CNN   & U & M \\
CROCS \cite{kiyasseh2021crocs}           & Comparative-based   & CNRV    & CNN   & S & I \\

LDVR \cite{anand2020unsupervised}        & Comparative-based   & CNRV    & CNN   & U & I \\
RDDC \cite{trosten2019recurrent}         & Comparative-based   & CLS    & RNN   & U & M\\
TSTCC \cite{eldele2021TSTCC}             & Comparative-based   & CNRV    & TRAN   & U & M\\
PCL \cite{Li2021PCL}                     & Comparative-based   & CNRV    & CNN   & U & M\\
CCL \cite{sharma2020clustering}          & Comparative-based   & CNRV    & CNN   & U & M\\
CRLI \cite{Ma_Chen_Li_Cottrell_2021}               & Comparative-based   & ADV    & GAN   & U & M\\
TS-CTS \cite{chang2022time}              & Comparative-based   & CNRV    & CNN   & Se & M\\
TNC \cite{tonekaboni2021unsupervised}    & Comparative-based   & CNRV    & CNN   & U & M\\
\addlinespace[0.1cm]
\hline
\addlinespace[0.1cm]
IDEC \cite{guo2017improved}     & Generative-based         & REC  & AE       & U  & I \\ 
DEPICT \cite{ghasedi2017deep}   & Generative-based         & REC  & AE       & U  & I \\
DCN \cite{yang2017towards}      & Generative-based         & REC  & AE       & U  & M \\
CKM \cite{gao2020deep}          & Generative-based         & REC  & AE       & U  & I \\
SOM-VAE \cite{fortuin2018deep}  & Generative-based         & REC  & VAE      & U  & I \\
SDCN  \cite{bo2020structural}   & Generative-based         & REC  & GCN      & U  & I \\

DTCR \cite{ma2019learning}      & Generative-based         & REC  & RNN      & U  & I \\
DTC \cite{madiraju2018deep}     & Generative-based         & REC  & LSTM     & U  & I \\
VADE \cite{jiang2017vade}& Generative-based         & ELBO & VAE  & U  & I \\
TST \cite{zerveas2021transformer} & Generative-based         & REC & TRAN  & U  & M \\
KMRL \cite{kashtalian2021k}     & Generative-based         & REC  & LSTM     & U  & I \\
T-DPSOM \cite{manduchi2019dpsom}& Generative-based         & REC  & LSTM     & U  & I \\ 
DTSS \cite{huang2022deep}       & Generative-based         & FCST  & TCN      & U  & M \\
DeTSEC \cite{ienco2020deep}     & Generative-based         & REC  & GRU      & U  & M \\
IT-TSC \cite{xu2021deep}        & Generative-based         & FCST  & TCN      & U  & M \\

\addlinespace[0.1cm]
\hline
\addlinespace[0.1cm]

DTCC \cite{zhong2023deep}      & Hybrid-based$^*$ & REC+CNRV &  LSTM & U  & I \\
TimeCLR \cite{yang2022timeclr} & Hybrid-based & REC+CNRV & CNN & U  & I \\
conDetSEC \cite{ienco2023deep} & Hybrid-based & REC+CNRV & CNN & Se  & M \\
MCAE \cite{xu2021unsupervised} & Hybrid-based & REC+CNRV & LSTM & U  & M \\
\hline
\addlinespace[0.1cm]
GPT4TS \cite{zhou2023onefitsall} & Foundation-Model & / & LLM & U & M \\
Chronos \cite{ansari2024chronos}  & Foundation-Model & FCST & LLM & U &  I\\
MOMENT \cite{goswami2024moment} & Foundation-Model & REC & TRAN & U &  I\\
TimesFM \cite{das2024timesfm} & Foundation-Model & FCST & LLM & U & M\\
UniTS \cite{gao2024units} & Foundation-Model & REC & * & U & M\\

\addlinespace[0.2cm]

\Xhline{2\arrayrulewidth} \addlinespace[0.2cm]
\end{tabular}
    \begin{tablenotes}
      \scriptsize
      \centering
      \item I: Univariate, M: Multivariate; Se: Semi-Supervised, U: Unsupervised; *: Arbitrary model backbone.
      \item Hybrid-based: Methods using learning strategies from both comparative-based and generative-based categories.
     \end{tablenotes}
\label{tab: representation-learning}
\vspace{-2em}
\end{table}%

\subsection{Time-series Representation-learning Overview}
\label{sec:represent_learning_overview}
Representation learning has been widely applied in numerous research topics, e.g., computer vision, natural language processing, etc. In early studies, researchers have found that while deep neural networks are mostly black boxes to human understanding, certain layers' output feature maps are capable of extracting meaningful feature information, such as edges or repeated patterns. In many cases, this kind of representation learned by a well-designed model architecture pre-trained on large datasets tends to be highly robust and could be easily adapted to downstream tasks. As the dimension of the learned representation can be highly reduced compared to the original data, these techniques have been widely adopted as a dimension reduction strategy in the time-series domain, where the input length could be extremely large. Figure \ref{pic:tsne} shows an example of the learned representation (visualized by T-SNE).

In general, given the learning strategy design, there are three different representation learning categories: supervised learning, semi-supervised learning, and unsupervised learning. In this survey, under the settings of the time-series clustering task, we mainly focus on unsupervised learning (some semi-supervised techniques are also included). The overall pipeline of the representation-learning-based methods in the clustering tasks can be summarized as follows: 

\begin{itemize}
    \item Pre-training stage. Given the designed model and training set, the objective of the pre-training stage is to learn a representative latent space suitable for time-series clustering. 
    \item Clustering stage. The representation obtained from the pre-trained model is utilized as a new input for conventional clustering methods, such as k-Means and k-Shape.
\end{itemize}

Following the discussion in \cite{lafabregue2021end}, there are three major components in time-series representation-learning-based clustering methods: (i) model architecture (ii) pretext loss (iii) clustering loss. The model architecture will influence the overall performance and inference speed, while the pretext loss and clustering loss design determine the structure of the latent space. Each component will be discussed in the following sections.

\subsubsection{Model Architecture} 

There are many different architecture designs in deep learning-based models, as well as in the time-series clustering domain. Among them, the most basic ones are Fully Connected Neural Network (FCN), Convolutional Neural Network (CNN), Recurrent Neural Network (RNN), Attention-based Neural Network, and Graph Neural Network (GNN). \\

\noindent\textbf{Fully Connected Neural Network}

\begin{figure}[!htp]
	\centering
	\includegraphics[width=0.85\linewidth]{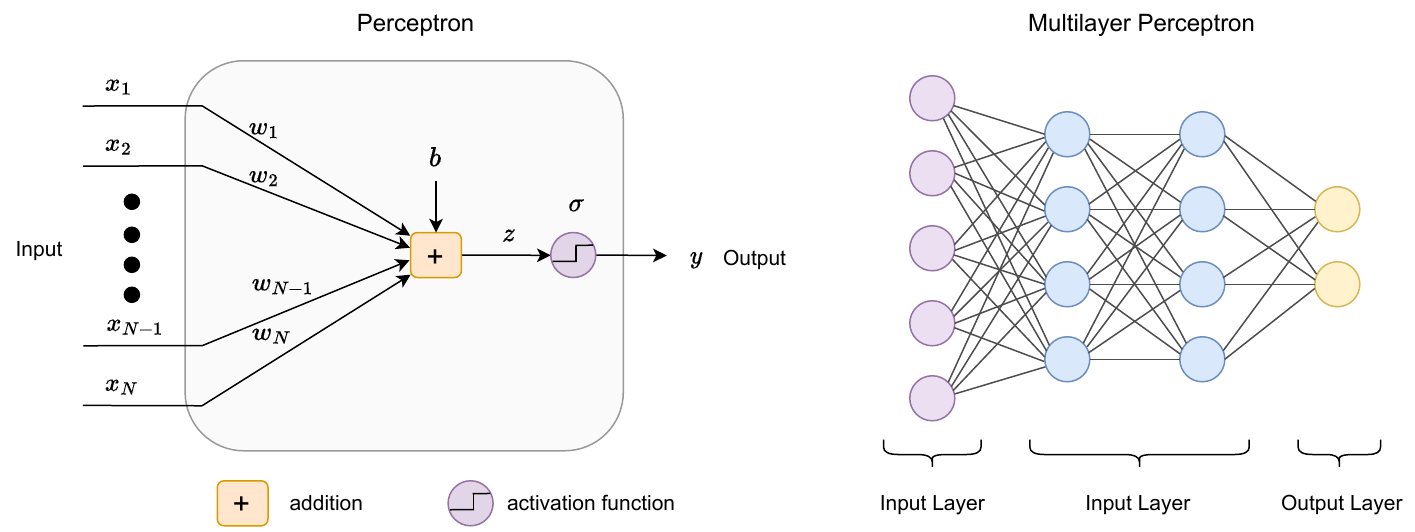}
	\caption{The overview of the structure of perceptron and the multilayer perceptron (MLP).}
	\label{pic:mlp}
\end{figure}
Fully Connected Neural Network (FCN), also known as Multilayer Perceptron (MLP), is one of the most basic architectures in deep learning-based models, originating from the idea in Neuroscience \cite{rosenblatt1958perceptron}. It proposes a hypothetical nervous system called a \textit{perceptron}. Each perceptron calculates a mapping between multiple inputs and one single output using learnable weights and activation functions (shown in Figure \ref{pic:mlp}). To extract different features, multiple perceptrons can be stacked together across layers and generate an output that usually possesses a lower dimension compared to the original input. Utilizing this idea, different FCN-based models have been proposed in various domains \cite{guo2017improved,xie2016unsupervised}
. Each layer of the network is named a dense layer, or a fully connected layer, consisting of multiple perceptrons as mentioned above. In mathematical form, the function of the $i$th layer can be expressed in the following equation:

\begin{align}
    Z_i &= W_i^TX_i+b_i,\ Y_i = \sigma(Z_i)
\end{align}
where $X_i$, $Z_i$, and $Y_i$ denote the input, intermediate, and output value of this current $i$th layer. $W$ and $b$ represent the learnable weights in the model. $\sigma$ here represents the activation function, which plays an important role in introducing the nonlinearity in neural networks. In practical use, the activation function $\sigma$ includes but is not limited to sigmoid, tanh, ReLU, and GELU.
Different choices usually depend on factors such as learning tasks, model structures, datasets.

\noindent\textbf{Convolutional Neural Network}

\begin{figure}[!htp]
	\centering
	  \begin{subfigure}[t]{0.45\linewidth}
        \centering
        \includegraphics[width=\linewidth]{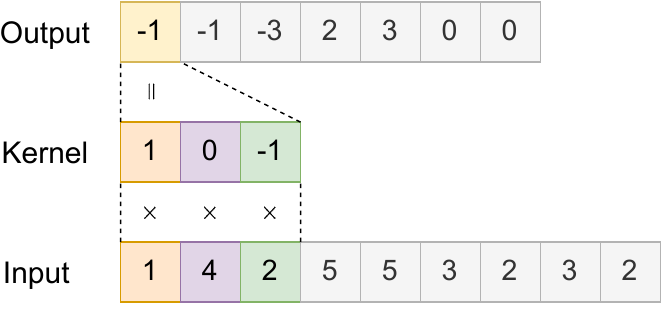}
        \caption{1D Convolution}
        \end{subfigure}
        \hspace{0.02\textwidth}
        \begin{subfigure}[t]{0.45\linewidth}
        \centering
        \includegraphics[width=\linewidth]{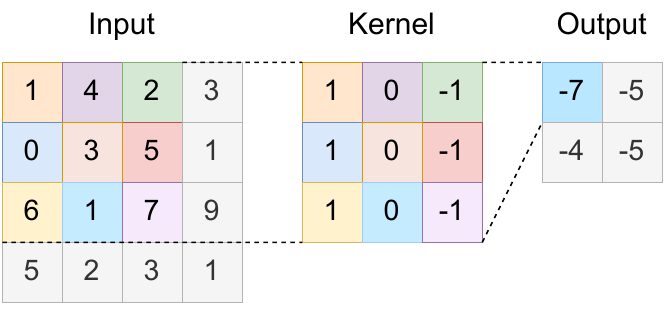}
        \caption{2D Convolution}
        \end{subfigure}
	\caption{The overview of the convolution mechanism.}
	\label{pic:cnn}
\end{figure}

Convolutional Neural Network (CNN) has been widely applied as the basic structure, especially in computer vision tasks. Representative CNN-based models are VGG \cite{simonyan2014very}, AlexNet\cite{krizhevsky2012imagenet}, and ResNet \cite{he2016deep}, which have achieved great success in different tasks such as image classification, detection, segmentation. As depicted in Figure \ref{pic:cnn}, each convolutional layer has $m$ \textit{kernels} of size $k$, which consist of learnable weights. Basically, each kernel can be viewed as a filter, scanning the data input in a sliding window way and searching for a certain pattern across dimensions. Compared with the FCN design, CNN layers take advantage of the kernel design which can be applied and reused across all dimensions, or channels, at the same time and thus need much fewer parameters than fully connected layers. Most widely used convolutional layers are in the format of 2-dimension (for image) or 1-dimension (for speech, text, and time-series data). \\

\noindent\textbf{Recurrent Neural Network}

However, an issue persists for both the fully connected neural network and the convolutional neural network discussed above: they are not able to accommodate inputs with varying lengths, an essential characteristic for some data formats such as speech, text, or general time series. To solve this problem, the recurrent neural network (RNN) is proposed \cite{hopfield1982neural,hochreiter1997long,cho2014learning}, which has become one of the most important model architectures in various research areas such as speech recognition, machine translation, and time-series clustering. 
Unlike traditional feedforward neural networks, the internal memory mechanism allows RNN-based models to take the input step by step and recursively update the hidden states (shown in Figure \ref{pic:lstm}). Representative models are Long Short Term Memory (LSTM) \cite{hochreiter1997long} and Gated Recurrent Units (GRU) \cite{cho2014learning}, which have achieved great success in different fields. In
mathematical form, the updating rule of the hidden state $h_t$ in traditional RNN can be expressed in the following equation (at time step $t$):
\begin{figure}[!htp]
	\centering
	\includegraphics[width=0.6\linewidth]{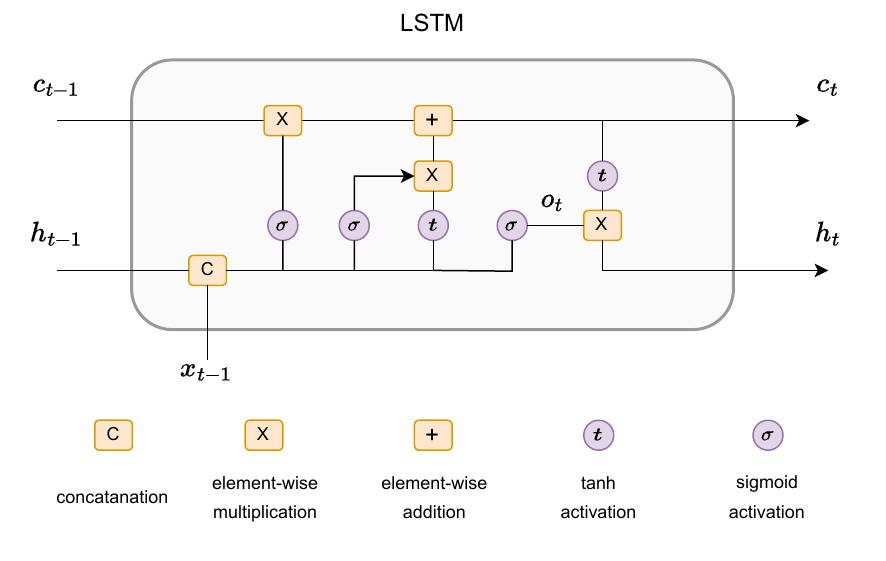}
	\caption{The overview of the LSTM architecture, one of the classic RNN-based networks.}
	\label{pic:lstm}
\end{figure}
\vspace{0.5em}
\begin{equation}
    h_t = \tanh(W_{h}h_{t-1}+W_xx_t+b),
\end{equation}
where $x_t$ and $h_t$ denote the input and hidden state at time step $t$. $W_h$, $W_x$ and $b$ represent the learnable weights in the RNN design. It is noted that RNN could also have multiple layers like FCN or CNN above. \\

\noindent\textbf{Attention-based Neural Network}

The concept of the attention mechanism, a widely recognized deep learning architecture, is proposed in the paper \cite{bahdanau2014neural}. It introduces the soft alignments between the encoder and decoder and achieves an apparent improvement in neural machine translation. In 2017, the Transformer architecture was introduced in the paper titled ``Attention is all you need" by Vaswani et al. \cite{vaswani2017attention}. This innovation quickly became the dominant model architecture in diverse research domains, including natural language processing (NLP), computer vision (CV), etc. 
It describes a mapping function from query, key, and value pairs to an output, where each component can be modeled from the input of each layer. The attention mechanism stands out for its good explainability and great performance, while greatly reducing the computational time with parallel computing. Transformer (TRAN), one of the most important attention-based architectures, is shown in Figure \ref{pic:transformer}. Give the query $Q$, key $K$ and value $V$, the attention mechanism can be expressed in the following way:

\begin{figure}
\centering
  
\includegraphics[width=0.7\linewidth]{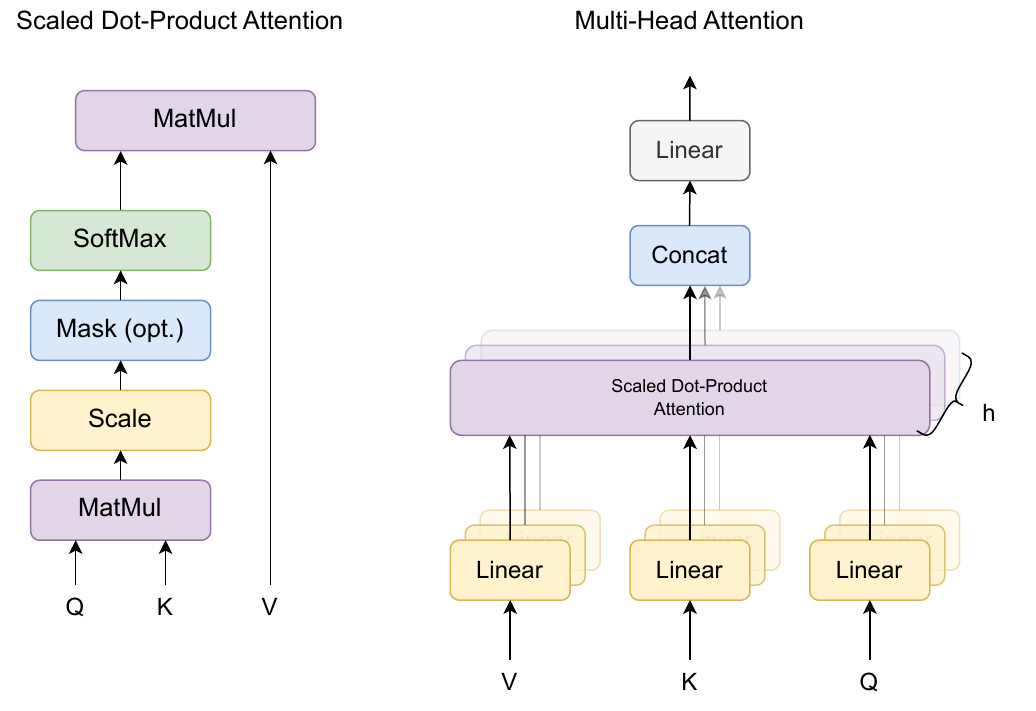}
    
\caption{The overview of the attention mechanism in Transformer \cite{vaswani2017attention}. }
\label{pic:transformer}
\end{figure}

\begin{equation}
    \text{Attention}(Q, K, V)=\text{softmax}(\frac{QK^T}{\sqrt{d_k}})V,
\end{equation}
where $\frac{1}{\sqrt{d_k}}$ is a scaling factor given the dimension of keys $d_k$. \\


\noindent\textbf{Graph Neural Network}

\begin{figure}[!htp]
	\centering
	\includegraphics[width=0.6\linewidth]{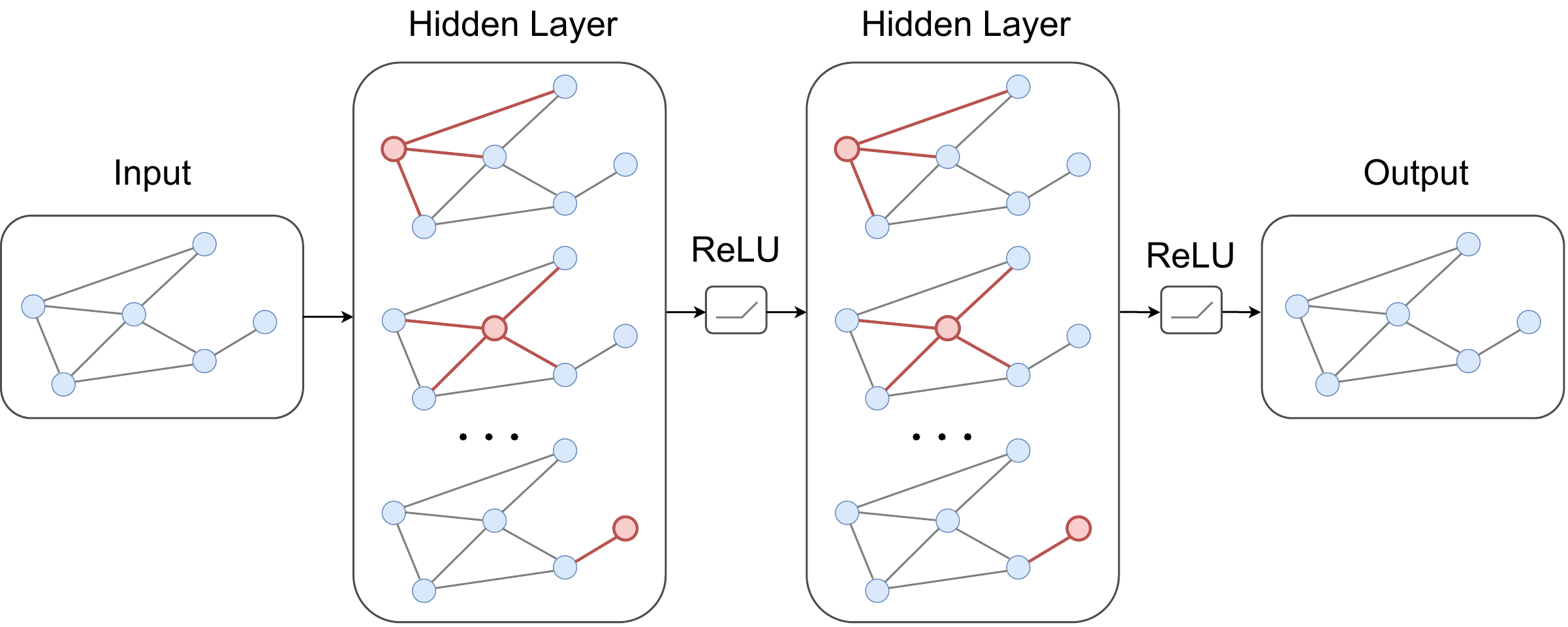}
	\caption{The structure of a simple graph neural network.}
	\label{pic:gnn}
\end{figure}

Graph, as one of the most important data structures, has been widely explored in many research fields. It can be seen as a collection of nodes/vertices and edges. Following the development of deep neural networks, graph neural network (GNN) receives great attention, particularly in scenarios where the relation between data samples are crucial for solving the problem. There are various GNN architectures, such as Graph Convolutional Networks (GCN) \cite{kipf2017semisupervised},  Graph Sample and Aggregated (GraphSAGE) \cite{hamilton2017graphsage}, Graph Attention Networks (GATs) \cite{veličković2018gats}. In time-series analysis, each data sample can be viewed as a node in a high-dimension space. The edges between nodes can be defined as the similarity or distance value. Considering the advantage of GNN in learning structural information, prior studies combine the GNN module with the traditional DNN module to obtain a better representation for time-series clustering \cite{bo2020structural}. In general form, the message and aggregation in each GNN layer can be expressed as Eq. \ref{eq:GNN}:

\begin{align}
    h_v^{(k)} &= f^{(k)} \left( W^{(k)}\cdot \frac{\sum_{u\in \mathcal{N}(v)} h_u^{(k-1)}}{|\mathcal{N}(v)|} + B^{(k)}\cdot h_v^{(k-1)}\right),
    \label{eq:GNN}
\end{align}
where $h_v^{(k)}$ represents the node embedding of $v\in V$ after $k^{th}$ layer. $\mN(v)$ denotes the neighborhood of the noce $v$. $W^{(k)}$ and $B^{(k)}$ are the learnable weights  and $f^{(k)}$ represents the activation function for $k^{th}$ layer. Figure \ref{pic:gnn} depicts the structure of a simple graph neural network.

\subsubsection{Pretext Loss}
Following the discussion in the prior study \cite{lafabregue2021end}, the objective functions of the representation learning in the first stage can be categorized into two different losses: one is pretext loss, which helps the model learn meaningful features. However, the pretext loss designs are usually targeted at general time-series analysis, not specifically clustering tasks. To solve this issue, many methods integrate clustering loss in the pre-training stage which provides a certain constraint manifold of the latent space. In this section, we are going to discuss the pretext loss.

\vspace{1em}
\noindent\textbf{Reconstruction Loss (REC)}

Reconstruction loss is one of the most widely used loss functions for general representation learning. By minimizing the error between the original input and the reconstructed output, the autoencoder could learn meaningful features that could be extracted for representing the data using much fewer dimensions.
\begin{align}
    \mL_{REC} &= \frac{1}{N} \sum_{i=1}^{N} \lVert X_i - \mD(\mE(X_i)) \rVert ^2,
    \label{eq:REC}
\end{align}
where $X_i$ is the $i$th input in the dataset, with $N$ samples in total. $\mE$ and $\mD$ are the encoder and decoder module of the autoencoder respectively. Usually the reconstruction loss takes L2 norm as shown in the equation above.

\vspace{1em}
\noindent\textbf{Multi-Reconstruction Loss (MREC)}

This is an extended version of the reconstruction loss, where the loss is calculated for each level. This hierarchical design puts stronger constraints on the representation learning. It is noted that the symmetry structure is required for the encoder and decoder model design.
\begin{align}
    \mL_{MREC} = \frac{1}{N} \sum_{i=1}^N\sum_{j=1}^{L} \lVert o^i_{\mD_j} - o^i_{\mE_j} \rVert ^2,
\end{align}
where $o^i_{\mE_j}$ and $o^i_{\mD_j}$ represent the output from the $j^th$ layer of the encoder and decoder.

\vspace{1em}
\noindent\textbf{Variational Autoencoder Loss (VAE)}

Variational autoencoder (VAE), as one of the most famous variants of the autoencoder (AE), is proposed to improve the capability of generalizing new data. A KL Divergence constraint is introduced to help regularize the probability distribution of the learned latent space compared to the pre-defined prior distribution (usually Gaussian distribution).
\begin{align}
    \mL_{VAE} &= \sum_{i=1}^N - \mathbb{E}_{z_i \sim q(z_i | x_i)} [\log p(x_i | z_i)] + KL (q(z_i|x_i)||p(z_i)),
\end{align}
where $z_i$ is the learned latent representation of the input data $x_i$, $p(z_i)$ is the pre-defined prior distribution, $q(z_i|x_i)$ and $p(x_i|z_i)$ are two conditioned distribution which could be modeled by deep neural networks.

\vspace{1em}
\noindent\textbf{Triplet Loss (TRPLT)}

Triplet loss is one of the most important objective function designs widely used in contrastive learning. Given the anchor data sample, we could define the positive and negative pair by carefully selecting similar/dissimilar data samples. The overall goal is to pull closer the learned representation of similar data samples and push apart those of dissimilar samples.
\begin{align}
    \mL_{TRPLT} &= -\log (\sigma(\mE(X)^{\top}\mE(X^+))) - \sum_{k=1}^{K}\log(\sigma(-\mE(X)^{\top}\mE(X^-_k))),
    \label{eq:triplet}
\end{align}
where the data sample $X$ is the anchor sample, $X^+$ and $X^-$ represent the positive and negative samples respectively. Denote by $\sigma$ the sigmoid function and $K$ the number of negative pairs.

\vspace{1em}
\noindent\textbf{InfoNCE Loss}

InfoNCE loss is another widely used contrastive learning loss design, widely used in unsupervised representation learning works such as SimCLR \cite{chen2020simple}. Similar to the triplet loss, the InfoNCE loss also enforces the model to capture the distance relationship between data pairs, e.g., positive pairs and negative pairs. Here we provide the equation in SimCLR \cite{chen2020simple} for illustration.
\begin{align}
    \mL_{InfoNCE} (i,j) &= -\log \frac{\exp(\text{sim}(z_i, z_j)/\tau)}{\sum_{k=1}^{2N} \mathds{1}_{[k\neq i]}\exp(\text{sim}(z_i, z_k)/\tau)} \label{eq:infonce}
\end{align}
where $z_i$ and $z_j$ are learned representations of $i^{th}$ and $j^{th}$ data samples in the dataset. $\text{sim}(\cdot)$ is the similarity function between two vectors. $\mathds{1}$ denotes the indicator function. $N$ is the batch size during training and $\tau$ is the temperature parameter.

\subsubsection{Clustering Loss}
As discussed in the previous section, pretext losses can be applied for general deep representation learning, but may not specifically deal with the issue in clustering tasks. Here we list 7 widely used clustering losses in the literature to help solve this problem.

\vspace{1em}
\noindent\textbf{DEC}

This loss is first proposed in DEC \cite{xie2016unsupervised} for unsupervised deep embedding learning. An auxiliary target distribution is introduced to help learn a representation suitable for clustering analysis. The loss is defined in Kullback–Leibler (KL) divergence format.
\begin{align}
    \mL_{DEC} &= \text{KL}(\mathbf{P}\lVert \mathbf{Q}),
\end{align}
where $\mathbf{P}$ represents the auxiliary target distribution and $\mathbf{Q}$ denotes the soft clustering assignment distribution. $N$ and $K$ are the batch size and number of clusters.

\vspace{1em}
\noindent\textbf{IDEC}

IDEC \cite{guo2017improved} is an extended version of DEC \cite{xie2016unsupervised}. an under-complete autoencoder is applied along with the clustering loss (DEC) for better local structure preservation. $\gamma$ is a hyperparameter to balance between the two losses. When $\gamma$ becomes 0, the IDEC loss will be the same as DEC loss.
\begin{align}
    \mL_{IDEC} &= (1-\gamma)\mL_{DEC} + \gamma\mL_{REC},
\end{align}

\vspace{1em}
\noindent\textbf{DEPICT}

Similar to IDEC loss, DEPICT \cite{ghasedi2017deep} also combines the clustering-oriented loss and the reconstruction loss. Different from the IDEC loss, the target distribution $\mathbf{P}$ also participates in the optimization process.
\begin{align}
    \mL_{DEPICT} &= \text{KL}(\mathbf{P} \Vert \mathbf{Q}) + \text{KL}(\mathbf{f} \Vert \mathbf{u}) + \mL_{MREC},
\end{align}
where $\mathbf{f}$ and $\mathbf{u}$ are the empirical label distribution and the uniform prior respectively. In the paper, the target distribution $\mathbf{P}$ is computed using the clean pathway design.

\vspace{1em}
\noindent\textbf{SDCN}

SDCN \cite{bo2020structural} utilizes a graph neural network to capture the structure information of the data. The overall objective function includes the reconstruction loss, clustering loss and GCN loss.

\begin{align}
    \mL_{SDCN} &= \mL_{REC} + \alpha\mL_{CLU} + \beta\mL_{GCN},
\end{align}
where $\mL_{CLU} = \text{KL}(\mathbf{P}\Vert\mathbf{Q})$ and $\mL_{GCN} = \text{KL}(\mathbf{P}\Vert\mathbf{Z})$. $\mathbf{Q}$ and $\mathbf{Z}$ are the soft cluster assignment distribution by the DNN and GCN modules respectively.

\vspace{1em}
\noindent\textbf{VaDE}

VaDE \cite{jiang2017vade} combines VAE and GMM to learn a deep representation that is well-suited for clustering tasks, and also capable of generating meaningful samples.
\begin{align}
    \mL_{VaDE} &= \mathbb{E}_{q(z,c|x)}[\log p(x|z)] - D_{KL}(q(z,c|x)\Vert p(z,c)),
\end{align}
where $c\sim \text{Cat}(\pi)$ denotes the cluster and $p(z,c)$ represents the Mixture-of-Gaussians (MoG) prior. The overall objective function is in the format of evidence lower bound (ELBO).

\vspace{1em}
\noindent\textbf{DTCR}

DTCR \cite{ma2019learning} incorporates an auxiliary classification module to discriminate between real and fake samples. Besides, a k-Means objective is also introduced to learn cluster-specific representations.
\begin{align}
    \mL_{DTCR} &= \mL_{REC} + \mL_{ADV} + \gamma \mL_{K-Means},
\end{align}
where $\mL_{ADV}$ resembles the discriminator in Generative adversarial network (GAN), which takes the format of cross-entropy loss during training.

\vspace{1em}
\noindent\textbf{ClusterGAN}

ClusterGAN \cite{ghasedi2019balanced} proposed a GAN-based model to help learn meaningful representations in an unsupervised manner. A clusterer module $\mC$ is introduced to help learn the mapping from the data sample to the latent space. 
\begin{align}
    \mL_{ClusterGAN} &= \text{min}_{\mG,\mC}\text{max}_{\mD} \mathbb{E}_{X\sim p(X)} [\log \mD(\mC(X), X)] + \mathbb{E}_{Z\sim p(Z)}[\log (1-\mD(Z,\mG(Z)))],
    \label{eq:ClusterGAN}
\end{align}

\subsection{Comparative-based}
\label{sec:comparative_based}
Similar to previous clustering methods, such as Encoding-based methods, the Comparative-based time-series clustering methods also provide a mapping function $\mathcal{E}: \mX \rightarrow \mZ$ to represent the time series in a latent vector (usually in a much fewer dimension) for downstream tasks. However, in this scenario, the encoder mapping function can be learned by a neural network in a comparative way, e.g., contrastive learning (CNRV) or generative adversarial networks (ADV). All methods are enumerated in Table \ref{tab: representation-learning}.






\subsubsection{Contrastive Learning}

\begin{wrapfigure}{r}{0.55\textwidth} 

	\centering
	\includegraphics[width=\linewidth]{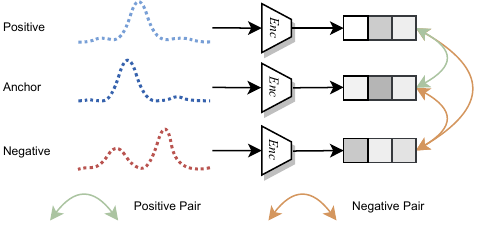}
	\caption{Overview of the contrastive learning strategy.}
	\label{pic:cnrv}
\end{wrapfigure}

Contrastive learning (CNRV) is one of the most widely-used techniques now in unsupervised deep learning, which has achieved great success in numerous research areas such as computer vision, natural language processing, speech recognition and so on. Given one data sample as the anchor, we could find similar and dissimilar examples to make: postive pairs and negative pairs. The goal of contrastive learning is to learn a robust representation such that the distances between postive pairs are shortened while the distance between negative pairs are enlarged (depicted in Figure \ref{pic:cnrv}). This strategy also naturally aligns with the core idea in some clustering techniques, e.g., partition-based methods, and thus has received great attention in recent years. In this section, we are going to introduce some widely-used contrastive learning loss and then go through the methods.

There are many different objective function designs for contrastive learning in general machine learning. Among them, two main objective functions or losses are widely used in time-series contrastive learning: Triplet Loss \cite{mikolov2013distributed,franceschi2019unsupervised}  and InfoNCE Loss \cite{oord2018representation,chen2020simple} (see Eq. \ref{eq:triplet} and \ref{eq:infonce}). Both of them have achieved great success in unsupervised representation learning and exhibit good performance in numerous downstream tasks.

Based on the number of negative samples $K$, we could assume one or multiple negative pairs in the experiments, while $\tau$ is a temperature parameter that controls the distribution \cite{wu2018unsupervised}. By carefully selecting positive and negative pairs, the model could learn a meaningful representation in a contrastive manner. Based on the view of contrast, we discuss the contrastive strategy from three different categories: Temporal Contrast, Instance-wise Contrast, and Multi-view Contrast. It is noted that the concept of Temporal Contrast and Instance-wise Contrast also clearly align with our two types of (whole) time-series clustering: clustering from the level of subsequences or the entire time series respectively.

\vspace{1em}
\noindent\textbf{Temporal Contast}

\noindent Temporal Contrast is first proposed in the \cite{franceschi2019unsupervised} to tackle the challenge of the positive/negative-pair selection under an unsupervised learning manner. With annotation, one could easily classify which should be positive or negative given the current anchor time-series sample. However, in unsupervised representation learning, such guidance is not available. Following word2vec's intuition, one solution is to use the idea of \textit{context}, to automatically generate positive and negative pairs. Given a random subsequence $x^{ref}$ in a time-series sample $x$, it is assumed that any subsequences $x^{pos}$ within the reference $x^{ref}$ should share a similar representation. On the other hand, any subsequence $x^{neg}$ obtained from other random time-series should be considered as negative samples with a large distance from $x^{ref}$. Triplet loss (T-Loss) is applied for the objective function. Numerous experiments demonstrate the learned representation exhibits good performance in the time-series clustering task.

Temporal Neighborhood Coding (TNC) \cite{tonekaboni2021unsupervised} adopts a similar concept as mentioned above. Instead of using T-Loss, a discriminator is applied to approximate the probability of one subsequence being the neighborhood of the other given the latent representation from the encoder. However, the aforementioned methods still do not take into account the selection of positive or negative pairs. In prior studies, it is found that anchors in minimal local variance could lead to poor performance \cite{chang2022time}. The author proposes a  Contrastive Triplet Selection strategy to select meaningful subsequence samples $x^{ref}, x^{pos}, x^{neg}$ based on the local variance and the similarity relationship calculated by Euclidean Distance, which provides more guidance to the model learning.

\vspace{1em}
\noindent \textbf{Instance-wise Contrast}

\noindent Instead of temporal information in a subsequence level, Instance-wise Contrast attempts to learn the similarity relationship from the level of instance, i.e., the entire time series. Given the positive pairs $\{x^{ref}, x^{pos}\}$ and negative pairs $\{x^{ref},x^{neg}\}$, the model can learn meaningful representation through contrastive learning. Here, all inputs, $x^{ref}, x^{pos}, x^{neg}$ are all distinct times-series from the given dataset, instead of subsequences. In order to generate reliable triplet sets, different strategies are explored. Ts2DEC \cite{ienco2019deep} proposes a semi-supervised deep embedding clustering frame that produces triplet constraints using ground truth, while methods like CCL \cite{sharma2020clustering} adopt the weak cluster labels from the learned representation.

Without supervision, the triplet set generation process faces the same situation as Temporal Contrast. As one of the earliest contrastive learning-based studies, SimCLR \cite{chen2020simple} explores a simple framework using image pairs generated by data augmentation. In the paper, two augmented views from the same data sample are considered as one positive pair, while different samples in one batch are noticed as negative pairs for each other. This strategy has achieved great success in different data modalities, including time series. LDVR \cite{anand2020unsupervised} applies the same concept and proposes an unsupervised triplet section strategy for time-series data. Instead of using the original format, 2-D time-series images are generated to utilize the knowledge from the pre-trained model in the computer vision field, e.g., ResNet \cite{he2016deep}. DCRLS \cite{xiao2023deep} extends the SimCLR idea in time-series domain with multi-layer similarity contrasting and achieves good performance. Self-distillation is adopted as a regularization in the framework for knowledge transfer. 

Similar to Temporal Contrast, some studies also focus on different sampling strategies for the triplet selection. SleepPriorCL \cite{zhang2021sleeppriorcl} proposes a knowledge-based positive mining strategy to tackle the sampling bias problem in contrastive learning. Positive pairs are re-defined within a minibatch using the pre-calculated dissimilarity. Experiments indicate the performance improvement in the sleep staging task compared with baseline methods.

\vspace{1em}
\noindent \textbf{Multi-view Contrast}

\noindent Apart from the previous two types, some recent studies also explore the combination of different views for robust representation learning. TS2Vec \cite{yue2022ts2vec} proposes a unified framework that utilizes both Instance-wise and Temporal contrast in a hierarchical way. TS-TCC \cite{eldele2021TSTCC} introduces the cross-view task for both temporal and contextual contrasting.
In addition to the time domain, some researchers also explore the possibility of the frequency domain. TF-C \cite{zhang2022self} introduces a new frequency encoder to ensure both time and frequency consistency between instance-level pairs.
Within the same year, BTSF \cite{yang2022unsupervised} adopted the concept of frequency consistency as well. For better capturing the information from both domains, the bilinear feature is optimized in a fusion-and-squeeze manner iteratively. 

Nevertheless, the aforementioned strategies mainly emphasize general representation learning through contrastive learning strategies, but may not specifically address the issue of clustering tasks. Cluster or prototype-based contrastive loss has been proposed to help improve the clustering quality \cite{meng2023mhccl,kiyasseh2021crocs, Li2021PCL}. Similar to Temporal Contrast or Instance-wise Contrast, the approaches enforce distinguishable representation between different clusters. As cluster-level contrast learning is hard to optimize on its own, it is usually treated as an auxiliary loss for a multi-view purpose in these papers.

\subsubsection{Generative Adversarial Network}

\begin{wrapfigure}{r}{0.55\textwidth} 

	\centering
	\includegraphics[width=\linewidth]{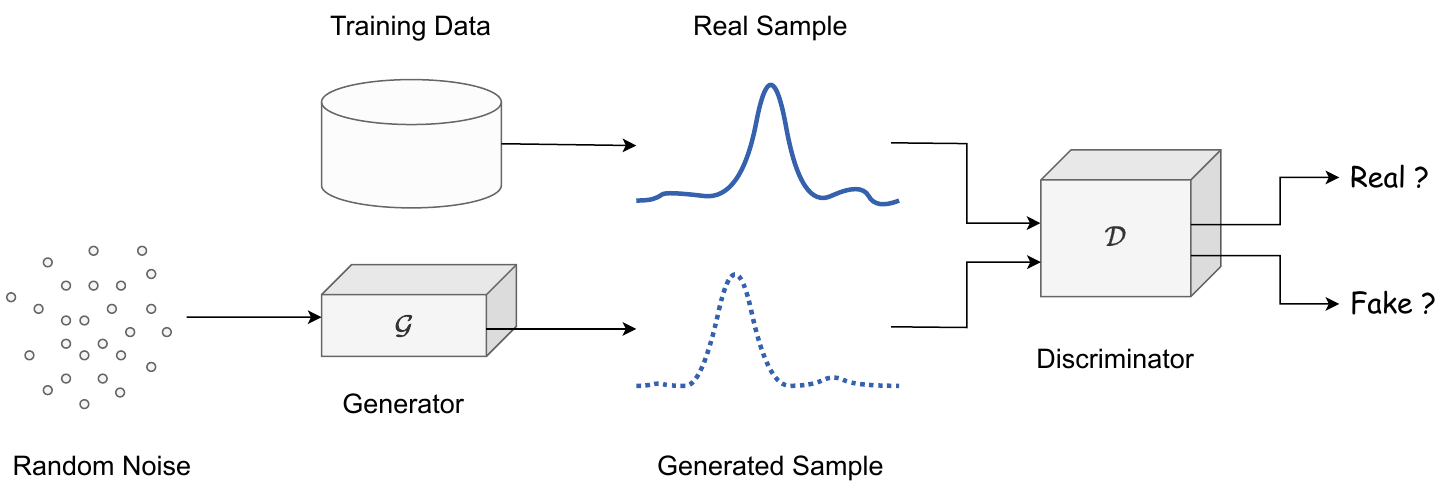}
	\caption{Overview of the generative adversarial networks (GAN).}
	\label{pic:gan}
    \vspace{-0.5em}
\end{wrapfigure}

Generative Adversarial Network (GAN) \cite{goodfellow2014GAN} is initially proposed for high-quality image generation, consisting of two major components: (i) the generative model $\mG$ attempts to capture the ground truth data distribution, and (ii) the discriminative model $\mD$ estimates the probability of the given sample being synthesized or real data. With the novel design of the minimax two-player game, the generator $\mG$ is capable of generating close-to-realistic data and achieved great success in all kinds of data formats including image, video, text, time series and so on. The structure of the classic GANs is shown in Figure \ref{pic:gan}.

Specifically, given a noise input $z \sim p_{z}(z)$, the generator will synthesize a fake data sample $\hat{x}=\mG(z)$, which shares the same shape as the real data $x\sim p_{data}$. The responsibility of the discriminator is to predict the probability that the given data $x^{\prime}$ is real, with a scalar of $\mD(x^{\prime})$, and $1 - \mD(x^{\prime})$ for synthesized data. In the end, the discriminator $\mD$ attempts to make a correct binary classification of the given data, and the generator $\mG$ will try to fool the discriminator $\mD$ with high-quality data samples. Denote the value function $V(\mD, \mG)$, the overall objective can be defined in a two-player minimax game way (Eq. \ref{eq:ClusterGAN}). The loss will finally converge when the Discriminator is unable to distinguish the synthetic data from the real data and the Generator could not improve itself as well ($\mD(x) = \mD(\mG(z)) = \frac{1}{2}$).

In order to distill the knowledge from GAN-based model, prior studies have explored different techniques. ClusterGAN \cite{ghasedi2019balanced} proposes a deep generative adversarial network for the clustering task. In order to capture the feature information of the data, ClusterGAN introduces a clusterer module that can map the real data into a discriminative representation. To address the problem in time-series domain, TCGAN \cite{huang2023tcgan} is introduced. Apart from the normal design of the generator and discriminator, a representation encoder is created using the pre-trained discriminator during the unsupervised learning process and thus tailored for downstream tasks such as time-series classification or clustering. Some studies also focus on specific cases of the time-series clustering task. For incomplete time-series clustering, CRLI \cite{Ma2021CRLI} provides an end-to-end method to optimize the imputation and clustering process using GAN-based network design. In this case, the generator can be viewed as an encoder that outputs the robust representation for future clustering processes.

\subsection{Generative-based}
\label{sec:generative_based}

Contrary to comparative-based clustering methods, generative-based time-series clustering methods utilize generative model architecture to learn the robust representation by casting constraints on the generation output. There are two major techniques in this field: (i) the reconstruction task: given input data, find a discriminative latent representation, which contains crucial feature information to reconstruct the original data. (ii) the forecasting task: given the subsequence of time-steps from 0 to $t-1$, predict the value of the next time step $t$. In order to accomplish this, an encoding process is also required for feature extraction and future prediction. In these ways, it is possible for us to find a good latent space for data representation with possibly fewer dimensions and we could just apply a simple K-means method on it to obtain the final clustering results. Methods can be found in Table \ref{tab: representation-learning}.

\subsubsection{Reconstruction-based Learning}
\begin{wrapfigure}{r}{0.55\textwidth} 
\centering
\includegraphics[width=\linewidth]{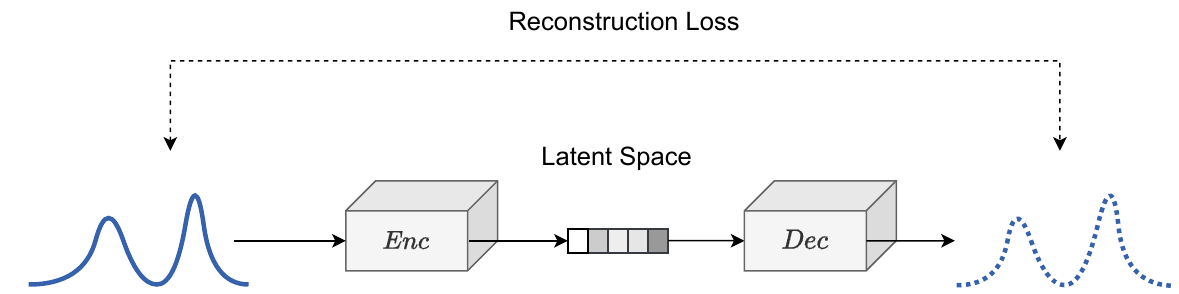}
\caption{Overview of reconstruction-based learning.}
\label{pic:rec}
\end{wrapfigure}

AutoEncoder (AE) is one of the most classic methods proposed for various tasks like dimension reduction, pre-training, and generation tasks using the concept of reconstruction \cite{kramer1991nonlinear,rumelhart1986learning,ballard1987modular,becker1991unsupervised}. Numerous variants, including Variational AutoEncoder (VAE) \cite{kingma2013auto}, have achieved great success in computer vision, machine translation, speech recognition and so on. Specifically, AutoEncoder consists of two important parts: an Encoder $\mE: \mX \rightarrow \mZ$, which maps the input $\mX$ to the latent space $\mZ$, and the Decoder $\mathcal{D}: \mZ \rightarrow \widehat{\mX}$ which produces reconstruction data $\widehat{\mX}$ from the latent space $\mZ$ (shown in Figure \ref{pic:rec}). The objective function can be defined as Eq. \ref{eq:REC}.

\vspace{1em}

Given the remarkable performance, there has been extensive investigation into AE-based time-series clustering techniques. Based on the observation that the classical method DEC \cite{xie2016unsupervised} could suffer from the distortion issue in the embedding space, IDEC  \cite{guo2017improved,madiraju2018deep} proposes to add the reconstruction constraint along with the clustering loss and better capture the structure information. DEPICT \cite{ghasedi2017deep} extends the idea and applies reconstruction loss at each level of the denoising autoencoder. Different components like the Noisy (corrupted) Encoder, Decoder, and clean Encoder are jointly optimized with the KL-divergence clustering loss.

Considering the wide usage of k-Means methods in representation-learning-based methods, many studies also investigate techniques to discover a ``k-Means friendly" space in reconstruction-based learning. For example, DCN \cite{yang2017towards} simultaneously optimizes the reconstruction and k-Means objective. However, the conventional k-Means loss design is non-differentiable and leads to the complexity of an alternating stochastic optimization strategy. To avoid that, CKM \cite{gao2020deep} introduces the deep k-Means strategies with concrete gradients and simultaneously optimizes the autoencoder parameters along with the cluster centroids. Following the concept of ``k-Means friendly" representation learning, some researchers also delve into the potential of incorporating other constraints on the latent space for better clustering. VaDE \cite{jiang2017vade} explores the combination of VAE and Gaussian Mixture Model (GMM). This integration helps learn good representations suitable for clustering tasks and enhances the capability to generate samples with Mixture-of-Gaussians (MoG) prior. Besides, other works also take the advantage of self-organizing map (SOM) design. SOM-VAE \cite{fortuin2018deep} and T-DPSOM \cite{manduchi2019dpsom} devise SOM-friendly representation learning frameworks by jointly optimizing the conventional VAE and the self-organizing map (SOM). This design strongly enhances the interpretability of time-series representation learning which benefits from the topological structure and smoothness in the learned latent space.
\begin{wrapfigure}{r}{0.5\textwidth} 
	\centering
	\includegraphics[width=\linewidth]{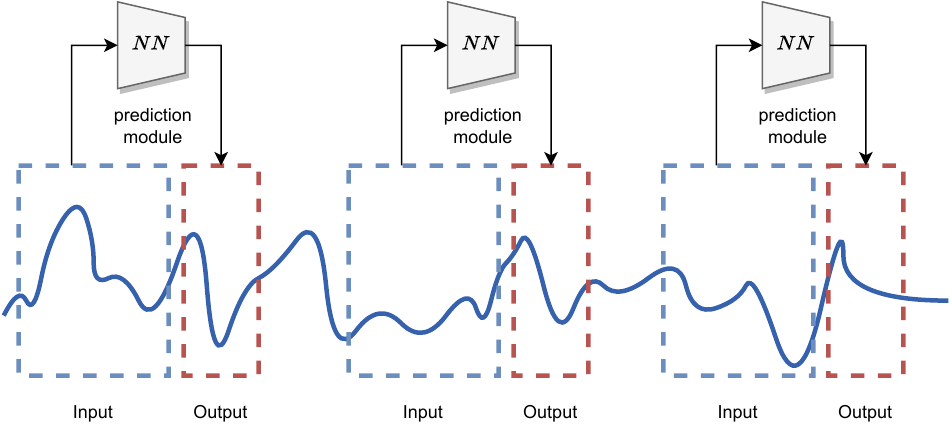}
	\caption{Overview of forecasting-based learning.}
	\label{pic:fcst}
\end{wrapfigure}

As various deep learning techniques continue to emerge, the influence of different model designs has become one of the focus in recent studies, e.g., GNN and attention-based neural networks. SDCN \cite{bo2020structural} proposes a dual self-supervised learning framework by incorporating the graph structure information from the GCN module. With this design, the knowledge can be transferred between the conventional autoencoder and the graph neural network for unified representation learning. As attention-based networks have become the state-of-the-art backbone design of many research fields, e.g., natural language processing, computer vision, etc., more and more researchers have been exploring its potential in time-series data representation learning. DeTSEC \cite{ienco2020deep} first introduces the attention and gating mechanisms in the conventional recurrent autoencoder. To learn an embedding manifold tailored for the clustering task, a clustering refinement process is applied in the second stage. TST \cite{zerveas2021transformer} proposes the first transformer-based network for multivariate time-series analysis. For unsupervised pre-training, the model is enforced to predict the value of masked segments. Extensive experiments have demonstrated the effectiveness of this modeling approach on diverse downstream time-series tasks.

\subsubsection{Forecasting-based Learning}
Similar to reconstruction-based learning, forecasting-based (FCST) learning models take the time-series data as an input and generate synthetic data as an output. Contrary to reconstruction-based learning, which learns the representation from reconstructing the data, forecasting-based learning focuses on predicting the next steps by learning the relationship between the past and future (as shown in Figure \ref{pic:fcst}). In one simple case of this kind of method, given a subsequence of time steps from 0 to $t-1$, the model predicts the value $\hat{x}_t$ at time step $t$. The regression loss will be calculated based on the error of predictions compared to the real data $x_t$ (Eq. \ref{eq:forcasting_loss}). It is noteworthy that depending on the forecasting scope, the regression objective function design could take different forms but the core idea remains consistent.

\begin{align}
    \mL_{FCT} &= \lVert \hat{x}_t - x_t \rVert
    \label{eq:forcasting_loss}
\end{align}

Some recent studies are delving into this learning strategy for time-series analysis. IT-TSC \cite{xu2021deep} devises a multi-path neural network to capture the variable association graphs, where each path is associated with one cluster. Given the subsequence $X(*, t_1 \colon t_{n-1})$, the model would autoregressively predict $\hat{X}(*, t_2 \colon t_{n})$. All the modules will be optimized together by minimizing the regression error. In the inference stage, the path with the least regression error will be assigned as the cluster label. DTSS \cite{huang2022deep} proposes a hybrid framework with temporal convolutional networks (TCN) and embedding sketching, which incorporates both local and global feature information. The embedding space is first trained by a forecasting-based learning strategy. For further dimension reduction, the sketch is extracted over the embedding space by sliding windows. Finally, the sequence of sketches is concatenated for clustering purpose.

\vspace{-1em}
\section{Evaluation Methods}
In this section, we discuss the evaluation algorithms for clustering, which serve as indices to assist individuals in evaluating the effectiveness of a clustering algorithm and deciding which one is suitable to use. Depending on whether they require external information or not, as discussed by \cite{aghabozorgi2015time}, clustering evaluation indices are generally divided into two types: external index and internal index. More specifically, the external index relies on external resources to assess the results of clustering, whereas the internal index evaluates the clustering based on the intrinsic structures of the results produced by clustering algorithms. In the following sections, we will delve into a comprehensive discussion of the concepts and representative evaluation methods associated with these two categories in Section \ref{sec:external_index} and \ref{sec:internal_index}.
\vspace{-1em}
\subsection{External Index}
\label{sec:external_index}
As described in \cite{aghabozorgi2015time}, the external index is employed to assess the resemblance between clusters generated by clustering algorithms and externally provided standards such as class labels and ground truth, making it the most popular method for evaluating clustering performance.

1. Purity: Obtaining the ground truth clusters and the generated clusters, in order to compute the purity of generated clusters with respect to the ground truth, each generated cluster is assigned to a class based on the majority class label within that cluster and then, the purity is calculated by dividing the total number of correctly assigned data points by the total number of data points.
\begin{equation}
    Purity = \frac{1}{N} \sum_{i} \max_{j} \left| C_i \cap T_j \right|
\end{equation}
where $N$ represents the total number of time-series data points, $C_i$ represents the cluster $i$, and $T_j$ represents for the ground truth class label $j$. The purity ranges from 0 to 1 and a poor clustering yields a purity approaching 0 while a perfect clustering achieves 1 as its purity. However, when cluster numbers are large and there are only a few number of time-series data points in each cluster, it is easy to achieve high purity. Considering the extreme situation that each cluster only contains one data point, the purity value is equal to 1. Hence, clustering quality evaluation cannot be performed solely relying on the purity.

2. Rand Index (RI): The RI was introduced by \cite{rand1971objective}. Given a set of $N$ data points $ \mX =\{x_0, x_1, \cdots, x_{N-1}\}$ and two clusterings generated on the same data $\mX$, the predicted clustering $C =\{C_0, C_1, \cdots, C_{K-1}\}$ and the ground truth $T = \{T_0, T_1, \cdots, T_{M-1}\}$, the RI assesses the similarity between $C$ and $T$ and it can be calculated by normalizing the number of similar assignments of point-pairs by the total number of point-pairs.
\begin{equation}
    RI = \frac{TP + TN}{TP + TN + FP + FN}
\end{equation}
where True Positives (TP) represents the frequency with which data points in a point-pair are grouped within the same cluster in both $C$ and $T$, True Negatives (TN) represents the frequency with which data points in a point-pair are assigned to different clusters in both $C$ and $T$, False Positives (FP) represents the count of occurrences where data points in a point-pair are clustered to the same cluster in $C$ while $T$ separated them into different clusters, and False Negatives (FN) represents the count of occurrences where data points in a point-pair are separated into the different clusters in $C$ while $T$ grouped them into the same cluster. The range of the RI value is from 0 to 1, where $RI = 1$ can be explained as two clusterings are identical and $RI = 0$ can be interpreted as two clusterings are completely distinct.

3. Adjusted Rand Index (ARI): The ARI was proposed by \cite{hubert1985comparing}, which is a corrected-for-chance extension of the RI. Given two random clusterings generated on the same set of data points, since the RI only considers the ratio of the number of similar assignments of point-pairs with respect to the total number of point-pairs, the RI may produce a relatively low or high value for this case. However, with corrections for chance, the ARI will generate value of 0 for random clustering results.
\begin{equation}
    ARI = \frac{RI - Expected \; RI}{Maximum \; RI - Expected \; RI}.
\end{equation}
The ARI generates values from -1 to 1, where $ARI = 1$ indicates that two clusterings are identical, $ARI > 0$ represents that the degree of similarity between two clusterings is better than random chance, $ARI = 0$ suggests that the agreement of two clusterings is equal to random chance, and $ARI < 0$ conveys that the agreement of two clusterings is worse than random chance in this case.

4. Normalized Mutual Information (NMI): NMI \cite{studholme1999overlap} is a measure widely employed in information theory and data analysis to evaluate the correlation or mutual information between two clusterings. Given two clusterings $M$ and $N$ on the same set of data points $X$, where $M$ and $N$ can have different numbers of clusters, NMI compares them by computing the mutual information between clusterings and normalizing it for the size of the clusters and the total number of data point. Based on \cite{studholme1999overlap} and \cite{strehl2002cluster}, NMI formula can be expressed as following equations:
\begin{align}
    NMI(M, N) &= \frac{I(M; N)}{\sqrt{H(M) \cdot H(N)}}, \\
    I(M; N) &= H(N) - H(N|M), \\
    H(M) &= -\sum_{m \in M} P(m) \log_2 P(m), \\
    H(N) &= -\sum_{n \in N} P(n) \log_2 P(n), \\
    H(N|M) &= -\sum_{m \in M, n \in N} P(m, n) \log_2\frac{P(m, n)}{P(m)}.
\end{align}
where $P(m, n)$ represents the joint probability of the value of $M$ is $m$ and the value of $N$ is $n$ while $P(m)$ and $P(n)$ represents the probability of $M$ taking on the value $m$ and the probability of $N$ taking on the value $n$ respectively. The score of NMI ranges from 0 to 1 where $NMI = 1$ implies the perfect correlation between two input clusterings and $NMI = 0$ suggests no mutual information between $M$ and $N$.

In addition to the methods mentioned earlier, there are several other notable external indices that can be used for clustering results evaluation. F-measure \cite{amigo2009comparison} combines both precision and recall into a unified value and is computed as the harmonic mean of these two metrics. Moreover, the F-measure ranges from 0 to 1, with the higher score suggesting better performance. The Entropy of a cluster \cite{amigo2009comparison} is another classic external index and evaluates the level of impurity among the data within each cluster. On top of that, entropy also ranges from 0 to 1, with the higher score indicating the higher uncertainty in the data of that cluster. The Adjusted Mutual Information (AMI) \cite{meilua2007comparing}, adjusted from the Mutual Information (MI), is designed to fix the fact that MI generates higher scores for two clusterings with more clusters, regardless of whether they share more information. Furthermore, AMI has a range from 0 to 1, where 1 represents an excellent agreement.



\vspace{-1em}
\subsection{Internal Index}
\label{sec:internal_index}
Also, as mentioned in \cite{aghabozorgi2015time}, the internal index is different from the external index and it is utilized to evaluate the quality and performance of a clustering structure without requiring external resources. 

1. Silhouette Coefficient: The silhouette coefficient \cite{rousseeuw1987silhouettes} is one of the classic metrics belonging to the internal index and is applied to measure the quality of clustering results obtained by clustering algorithms without requiring any ground truth. Given a single data point i in a dataset, its silhouette coefficient $s(i)$ can be calculated as follows:
\begin{equation}
    s(i) = \frac{b(i) - a(i)}{\max\{a(i), b(i)\}}
\end{equation}

where $b(i)$ is the minimum average distance from data point $i$ to data points in a different cluster and $a(i)$ is the average distance from data point $i$ to the other data points within the same cluster. For data point $i$, its $s(i)$ ranges from -1 to 1 where a large value closing to 1 indicates that data point $i$ is well-clustered while a small value approaching -1 suggests an opposite meaning. Based on the individual silhouette coefficient, the overall silhouette score can be calculated by averaging silhouette coefficients across all data points. A high silhouette score suggests that data points are well-clustered while a really low score indicates that the clustering is inaccurate. 

2. Davies-Bouldin index (DB index): DB index \cite{davies1979cluster}, which considers both dispersion within clusters and separation between clusters to measure the quality of clustering, is another representative and noteworthy clustering evaluation method that belongs to the internal index. The dispersion $S(i)$ of a cluster $C_i$ is calculated by the average distance between each point $x \in C_i$ and its centroid $c_i$, given by the following equation:
\begin{equation}
    S(i) = \frac{1}{\lvert C_i \rvert} \sum_{x \in C_i} d(x, c_i)
\end{equation}
Besides that, the separation between clusters $C_i$ and $C_j$, as exhibited by the following equation, can be measured through calculating the distance between centroids $c_i$ and $c_j$ of cluster $C_i$ and $C_j$:
\begin{equation}
     M(i, j) = d(c_i, c_j)
\end{equation}
After obtaining both dispersion within clusters and cluster separations, we are able to calculate the DB index for a clustering result that generates $K$ clusters through the function below:
\begin{equation}
    DB = \frac{1}{K} \sum_{i=1}^{K} \max_{j=1 \cdots K, j \neq i} \frac{S(i) + S(j)}{M(i, j)}
\end{equation}
For a clustering result assessed using the DB index, a lower score indicates that each cluster has a high internal similarity and is well-distinguished from other clusters.

3. Dunn index (DI): DI \cite{dunn1974well} is a classic internal index and its computation involves the consideration of both intra-cluster compactness and inter-cluster separation. To compute DI for $K$ clusters:
\begin{align}
    DI &= \frac{\min_{i=1,\cdots,K, j=1,\cdots,K, i \neq j} \{dist(C_i, C_j)\}}{\max_{p=1,\cdots,K} \{diam(C_p)\}} \\
    dist(C_i, C_j) &= \min_{x \in C_i, y \in C_j} || x - y || \\
    diam(C_p) &= \max_{x, y \in C_p} || x - y ||
\end{align}
where the cluster $i$ is denoted as $C_i$. If a clustering result has a high DI score, we can state that there exists compact and well-separated clusters. However, it is worth mentioning that DI is sensitive to outliers and computationally expensive.

4. Within-cluster sum of squares (WCSS): WCSS \cite{macqueen1967some}, a noteworthy internal index, evaluates the cluster cohesion which gauges the degree of similarity among objects within a cluster. Moreover, WCSS can be calculated by summing up the squared distances between each data $x$ and the centroid of its assigned cluster:
\begin{equation}
    WCSS = \sum_{i=1}^{K} \sum_{x \in C_i} d(x, c_i)^2
\end{equation}
where $K$ is the number of clusters and $c_i$ represents the centroid of the $i^{th}$ cluster $C_i$. Beyond that, WCSS is applied in the elbow method to determine the optimal number of clusters $K^*$ for clustering algorithms that require $K^*$ as an input. Evaluating a given clustering result by WCSS, a low score represents a desirable situation where each cluster's data are tightly grouped around the centroid. However, it is worth mentioning that an extremely low WCSS may be achieved by choosing a very large $K$ and, if we make $K$ equal to the number of data, WCSS will equal $0$.

\vspace{-1em}
\section{Conclusion}
Time-series clustering is an unsupervised task that aggregates similar time-series data into groups, aiming to reduce the intra-class distance and maximize inter-class distance. In this survey, we collect more than 100 time-series clustering algorithms, dividing them into 4 first-level categories: Distance-based, Distribution-based, Subsequence-based, and Representation-learning-based. To further classify the gathered time-series clustering algorithms, we propose 10 secondary-level categories that stem from their respective parent categories. 
Based on the proposed taxonomy,
we offer an in-depth discussion of key algorithms within each category. In addition, building on the prior studies \cite{aghabozorgi2015time}, we also investigate the external and internal indices utilized for evaluating the clustering results.

Despite decades of progress in this area, the challenge of time-series clustering still persists. Different clustering designs become crucial when the distortion of time series cannot be trivially diminished with pre-processing strategies in real-world scenarios \cite{paparrizos2015k,yang2011patterns,zakaria2012clustering,yue2022ts2vec}. 
Partitional clustering methods, starting from the classic k-Means, have demonstrated a good balance in clustering accuracy and runtime \cite{paparrizos2015k,javed2020benchmark}. Hierarchical clustering methods, in comparison, provide more flexibility in clustering resolution \cite{kakizawa1998discrimination,luo2023time,javed2020benchmark}, e.g., the dendrogram could be cut at different heights to obtain a finer or coarser clustering. As illustrated in previous sections, the choice of dissimilarity measure and representation plays an important role in determining the accuracy and runtime of one clustering method. Numerous approaches have been proposed to integrate new insights into these two crucial components from different perspectives, e.g., distribution-based and subsequence-based methods. The former focuses on modeling the distribution of the time-series data, such as the hidden Markov model (HMM) \cite{oates1999clustering,li2001building} of the training data or the density across the raw time-series space \cite{ester1996density,chen2007density}. On the other hand, the latter exploits representative subsequences to represent each time series and exhibit robustness to noise perturbations, as it mainly focuses on salient patterns \cite{zakaria2012clustering,ren2017sliding}. With the development of deep learning techniques, many representation-learning-based methods have been introduced and demonstrated effectiveness in this domain \cite{yue2022ts2vec,meng2023mhccl,franceschi2019unsupervised,lafabregue2021end}. These unsupervised learning approaches demonstrate strong performance in both dimensionality reduction and representational capability \cite{yue2022ts2vec,lafabregue2021end}.

To reveal the landscape in this domain, a few evaluation studies have emerged and received increasing attention. 
\cite{javed2020benchmark} provides the first time-series clustering benchmark with 8 popular methods from partitional, hierarchical, and density-based categories. A steady increase has been observed for newly proposed clustering approaches, however, no single method could outperform others in all datasets. 
\cite{lafabregue2021end} presents a comprehensive study on the effectiveness across model architecture, learning strategies, and parameter setting in deep learning-based clustering methods, which sheds light on this direction. 
\cite{paparrizos2023odyssey} provides a modular web engine named Odyssey that enables rigorous evaluation studies across 128 time-series datasets. 
Overall, there is no single method that proves superior across all scenarios, which highlights the need for in-depth investigation across various domains in future studies. A key issue is to find the right balance between clustering accuracy and runtime cost across various scenarios. We aim for this survey work to serve as a comprehensive exploration of this area, offering insights for future time-series clustering algorithm designs.



\bibliographystyle{ACM-Reference-Format}
\bibliography{reference/reference_update}

\end{document}